\documentclass{article}[final] % For LaTeX2e
\usepackage{iclr2026_conference,times}

    \PassOptionsToPackage{numbers}{natbib}
\PassOptionsToPackage{dvipsnames,svgnames,x11names}{xcolor}

\pdfoutput=1  %for arXiv, flag to use pdfLatex

\usepackage{amsmath,amsfonts,bm}

\def\eqref#1{equation~\ref{#1}}
\def\1{\bm{1}}

\DeclareMathAlphabet{\mathsfit}{\encodingdefault}{\sfdefault}{m}{sl}
\SetMathAlphabet{\mathsfit}{bold}{\encodingdefault}{\sfdefault}{bx}{n}

\newcommand{\E}{\mathbb{E}}

\usepackage[english]{babel}
\usepackage{hyperref}
\usepackage{url}
\usepackage{algorithm, algpseudocode}
\usepackage{graphicx} 
\usepackage[english]{babel}
\usepackage[utf8]{inputenc} % allow utf-8 input
\usepackage[T1]{fontenc}    % use 8-bit T1 fonts
\usepackage{booktabs}       % professional-quality tables
\usepackage{amsfonts}       % blackboard math symbols
\usepackage{nicefrac}       % compact symbols for 1/2, etc.
\usepackage{microtype}      % microtypography
\usepackage{todonotes}
\usepackage{amsmath}
\usepackage{placeins}
\usepackage{amsthm}
\usepackage{array} 
\usepackage{enumitem}
\usepackage{subcaption}
\usepackage{dsfont}
\usepackage[export]{adjustbox}
\usepackage{multirow}
\usepackage{cleveref}
\usepackage{xcolor}

\title{Unconditional CNN denoisers contain sparse semantic representations of images}
\author{Zahra Kadkhodaie \\
New York University \\
Flatiron Institute, Simons Foundation \\
\texttt{zk388@nyu.edu}
\And
Stéphane Mallat \\
Collège de France \\
Flatiron Institute, Simons Foundation \\
\texttt{stephane.mallat@ens.fr}
\And
Eero P.~Simoncelli \\
New York University \\
Flatiron Institute, Simons Foundation\hspace*{0.35in} \\
\texttt{eero.simoncelli@nyu.edu}
}

\newcommand{\sm}[1]{\textcolor{blue}{#1}}
\definecolor{mypink1}{rgb}{0.858, 0.188, 0.478}

\definecolor{escolor}{rgb}{0.1, 0.6, 0.0}

\long\def\myComment#1{}

\iclrfinalcopy % Uncomment for camera-ready version, but NOT for submission.

\begin{document}

\maketitle

%%%%%%%%%%%%%%%%%%%%%%%%%%%%%%%%%%%%%%%%%%%%%%%%%%%%%%%%%%%%%%%%%%%%%%%%%
%%%%%%%%%%%%%%%%%%%%%%%%%%%%%%%%%%%%%%%%%%%%%%%%%%%%%%%%%%%%%%%%%%%%%%%%%

\begin{abstract}
Generative diffusion models learn probability densities over diverse image datasets by estimating the score with a neural network trained to remove noise. Despite their remarkable success in generating high-quality images, the internal mechanisms of the underlying score networks are not well understood. Here, we examine the image representation that arises from score estimation in a {fully-convolutional unconditional UNet}. 
We show that the middle block of the UNet decomposes individual images into sparse subsets of active channels, and that the vector of spatial averages of these channels can provide a nonlinear representation of the underlying clean images. Euclidean distances in this representation space are semantically meaningful, even though no conditioning information is provided during training. We develop a novel algorithm for stochastic reconstruction of images conditioned on this representation: The synthesis using the unconditional model is "self-guided" by the representation extracted from that very same model. For a given representation, the common patterns in the set of reconstructed samples reveal the features captured in the middle block of the UNet. Together, these results show, for the first time, that a measure of semantic similarity emerges, \emph{unsupervised}, solely from the denoising objective.

\end{abstract}

%%%%%%%%%%%%%%%%%%%%%%%%%%%%%%%%%%%%%%%%%%%%%%%%%%%%%%%%%%%%%%%%%%%%%%%%%
%%%%%%%%%%%%%%%%%%%%%%%%%%%%%%%%%%%%%%%%%%%%%%%%%%%%%%%%%%%%%%%%%%%%%%%%%

% $x_\sigma = x + z$
% $\hat{f}(x_\sigma)$
% $\mathbb{E} \left [ \Vert x - f(x_\sigma)\Vert^2 \right] $\\
% \\

% $f(x_\sigma) = \mathbb{E}_x[x|x_\sigma] = \int xp(x|x_\sigma)dx$
% \\

% $\hat{f}(x_\sigma)= x_\sigma + \sigma^2 \nabla_{x_{\sigma}} \log \hat{p}_\sigma(x_\sigma)$
% \\

% $ \nabla_{x_{\sigma}} \log \hat{p}_\sigma(x_\sigma) = (\hat{f}(x_\sigma) -  x_\sigma )/\sigma^2 $
% \\

% $p_\sigma(x_\sigma)=  \int p(x_\sigma , x) dx = \int p(x_\sigma | x) p(x) dx = \int g_\sigma(x_\sigma - x) p(x) dx$

\section{Introduction}
Generative diffusion methods provide a powerful framework for sampling from probability densities learned from complex high-dimensional data such as images \citep{sohlDickstein15, song2019generative, ho2020denoising}. The key to their success lies in  deep neural networks trained to estimate the score (the gradient of the log of the noisy image distribution), which is achieved by optimization on a denoising task. % \citep{Robbins1956Empirical, Miyasawa61, hyvarinen2005estimation}.
Training and sampling algorithms for these models have been extensively studied \citep{surveyDiffuionVision}, but the
score estimation properties that enable their spectacular image generation capabilities are not understood. 
\myComment{\sm{Estimating the score is equivalent to computing an minimum mean-square denoising estimator.} }
Since denoising generally relies on distinguishing signal from noise, these score networks must somehow learn to identify and isolate patterns and structures found in their training data, so as to preserve them while eliminating noise.
Here, we open the ``black box'' of a score network  to reveal this internal representation. 
%\sm{It is an important step to build models of how deep networks achieve a nearly minimum mean-square denoising of images.}

% The majority of generative diffusion models use denoisers that are trained {\em conditionally}: along with a noise-corrupted image, the network receives auxiliary information, most commonly derived from text instructions \citep[][e.g.,]{??}, or from a recognition network \citep[][e.g.,]{??}. 
Here, we examine a UNet \citep{ronneberger2015u}, trained \emph{unconditionally} for image denoising. Prior work \citep{Brempong_2022_CVPR, yang2023diffusion, xiang2023denoising, baranchuklabel, mukhopadhyay2023diffusion} has shown that activations from such a model can be extracted to perform downstream tasks -- such as classification and segmentation -- with considerable success.
We aim instead to understand and interpret those aspects of the internal representations of a denoiser that arise solely from the denoising task. 
We demonstrate that this network computes a low dimensional \emph{sparse} set of activations in the output channels of the middle network block. This may be summarized with a vector comprised of the spatial averages of these channels, which provides a representation of the underlying clean image. We verify that this representation is stable when computed on an image contaminated with noise over a broad range of amplitudes. 
% consistent with the denoising objective on which it is trained. 
%\sm{because it computes a denoiser which aims at eliminating the influence of the noise.}

This representation exhibits a number of intriguing properties. Roughly, the channels summarized by the representation vector fall into two categories: non-selective channels, which capture common features present in many images; and selective channels, which are specialized for patterns that occur in only a small subset of images. As a consequence, for diverse datasets, the representation vectors lie within a union of low-dimensional subspaces, each of which is spanned by many of the common channels and a small subset of the specialized channels.
We demonstrate that distances in this space are meaningful: Images whose representation vectors are similar are also semantically similar. This allows us to partition the images in the dataset by applying a clustering algorithm to their representation vectors. The emergent clusters capture the general visual appearance of the corresponding images, sharing both fine details as well as global structure, but are only partially aligned with object category labels. 

Finally, we develop an algorithm for stochastic sampling, using a reverse diffusion algorithm conditioned on this representation vector computed from a target image. This procedure recovers a sample from a set of images whose representation is the same as that of the target image. 
Visually, these images are similar in terms of both local and global patterns, revealing both the commonality and diversity of attributes encoded in the representation.
To quantify this, we show that
the Euclidean distance between a pair of representation vectors strongly predicts the distance between a pair of conditional distributions induced by the representations. 
Thus, we show that the denoising objective alone, without any external conditioning, engenders learning of high level features that carry detailed semantic information.

%%%%%%%%%%%%%%%%%%%%%%%%%%%%%%%%%%%%%%%%%%%%%%%%%%%%%%%%%%%%%%%%%%%%%%%%%
%%%%%%%%%%%%%%%%%%%%%%%%%%%%%%%%%%%%%%%%%%%%%%%%%%%%%%%%%%%%%%%%%%%%%%%%%
\section{Implicit sparse image representation}
Diffusion models learn image densities from data using a network that is trained for denoising.  Specifically, given a noisy image $x_\sigma = x + \sigma z$, with $z\in \mathcal{N}(0,{\rm Id})$ a sample of white Gaussian noise, one trains a network $s_\theta(x_\sigma)$ to minimize the squared error:
\begin{equation}
 \ell(\theta) =  \mathbb{E}_{x,\sigma, z} \Vert x -  \hat{x}({x_{\sigma})} \Vert^2
 =  \mathbb{E}_{x,\sigma, z} \Vert  {s}_\theta(x_{\sigma}) - \sigma z \Vert^2 .
\label{denoising-loss}
\end{equation}
The optimal solution is the conditional mean, which can be directly related to the score of the underlying noisy distributions using a relationship published in \cite{Miyasawa61}, but generally referred to as "Tweedie's formula" \citep{Robbins1956Empirical,efron2011tweedie}:
\begin{equation}
    \hat{x}(x_{\sigma}) = \mathbb{E}[x|x_{\sigma}] =  x_{\sigma} + \sigma^2 \nabla_{x_{\sigma}} \log p_{\sigma}(x_{\sigma}) .
    \label{Miyasawa}
\end{equation}
Thus, the trained network provides an approximation of the family of score functions for all $\sigma$.  Reverse diffusion methods draw samples through iterative partial application of the learned denoiser, thereby using this approximate score to ascend the probability landscape
\citep{sohlDickstein15,song2020score,ho2020denoising,kadkhodaie2020solving}.
% kadkhodaie2020solving,
\myComment{%original text:
by estimating the family of noise-corrupted score functions, $\nabla_{x_{\sigma}} \log p_\sigma (x_{\sigma})$, where $p_\sigma$ is the density of $x_\sigma = x + \sigma z$, and $z\in \mathcal{N}(0,{\rm Id})$ is a sample of white Gaussian noise, for a wide range of noise amplitudes, $\sigma>0$. The score is estimated with parameterized a neural network $s_{\theta}(x_\sigma)$, and is learned by minimizing a denoising objective, 
\begin{equation}
 \ell(\theta) =  \mathbb{E}_{x,\sigma, z} \Vert x -  \hat{x}({x_{\sigma})} \Vert^2
 =  \mathbb{E}_{x,\sigma, z} \Vert  {s}_\theta(x_{\sigma}) - \sigma z \Vert^2 .
\label{denoising-loss}
\end{equation}
Indeed, it has been proven \citep{Miyasawa61, Robbins1956Empirical}
that the minimum mean-square error estimator (MMSE) which minimises this loss is specified by the score:
\begin{equation}
    \hat{x}(x_{\sigma}) = \mathbb{E}[x|x_{\sigma}] =  x_{\sigma} + \sigma^2 \nabla_{x_{\sigma}} \log p_{\sigma}(x_{\sigma}) .
    \label{Miyasawa}
\end{equation}
Reverse diffusion algorithms employ these scores to draw samples from $p(x)$ \citep{hyvarinen2005estimation,raphan2011least,vincent2011connection,song2019generative,ho2020denoising,sohlDickstein15,kadkhodaie2020solving}. 
}

%%%%%%%%%%%%%%%%%%%%%%%%%%%%%%%%%%%%%%%%%%%%%%%%%%%%%%%%%%%%%%%%%%%%%%%%%
\subsection{Spatial averages of activations}
We adopt a convolutional UNet architecture \citep{ronneberger2015u}, a smaller and more readily analyzed architecture than the most recent implementations, that nevertheless offers strong denoising performance. The model consists of three main components: a set of encoder blocks (the downsampling path), a middle block, and a set of decoder  blocks (the upsampling path) (see Appendix \ref{app:architetcure,training,dataset} for details). We examine the activations of these hidden layers to understand how the noisy image is transformed to produce the score. 
To reduce the dimensionality, we consider the simplest summary of channel activations, $a_j(x_\sigma)$, consisting of the spatial averages 
of all channels at the output layer of each block, notated by $\bar{a}_j(x_\sigma) \in \mathbb{R}^{d_j}$. The components of the vector $\bar{a}_j$ are nonnegative (due to half-wave rectification (ReLU) nonlinearities), and carry information about the features encoded by their corresponding channels. Because of the convolutional nature of the network, the activations within each channel indicate the presence or absence of a feature in different spatial locations. Thus if a feature associated with channel $i\in(0,d_j)$ is present somewhere in the input image, $\bar{a}_j(x_{\sigma})[i]$ will be positive, otherwise zero.

It is worth noting that the structure of the UNet architecture imposes certain properties on $\bar{a}_j$. Idealized convolutions are translation equivariant, hence the spatial averages of their activations are translation {\em invariant} w.r.t to the extracted features and do not carry information about location. 
For ergodic processes, these averages are approximations of statistical moments (expected value of nonlinear functions), and such measurements have been used for representation of visual texture \cite{julesz1962visual,zhu1998filters,portilla2000parametric,victor2017textures}. 
\myComment{
  \sm{probability distribution is essentially captured by a low-dimensional sparse representation which is similar to moments and that would be close to moment values for ergodic processes (textures) because the spatial averaging then provides an empirical estimation of moments. 
  Of course this is just a first step because we do not understand fully this representation and the properties of the encoder and decoder but the existence of this representation provides a way to approach the problem. }
  The relation between $\phi$ and moments gives an important intuition of what it corresponds to in simpler cases (textures).} 
However, translation equivariance is imperfect for UNets (and most other convolutional networks) since it is violated by zero-padded boundary handling, and downsampling operations. Both effects are more prominent in the deeper blocks of the encoder, which have undergone more downsampling, and for which the boundary influence encroaches on a larger spatial portion of the channels. In these cases, $\bar{a}_j$ can carry information about the location of features in addition to their presence. This effect coincides with the growth of the receptive field (RF) in the deeper blocks.  As a result, $\bar{a}_j$ is expected to carry location information for larger features \citep{kadkhodaie2023learning, kamb2024analytic}. 

%%%%%%%%%%%%%%%%%%%%%%%%%%%%%%%%%%%%%%%%%%%%%%%%%%%%%%%%%%%%%%%%%%%%%%%%%%%%%%%%%%%%%%
\subsection{Denoising and channel sparsity} 
One of the difficulties in studying image representation in diffusion models is that it is not obvious where to look. We aim to locate the layers whose feature vectors, $\bar{a}$'s, well represent the clean image underlying the noisy input image. Such feature vectors would be rich enough to capture patterns and regularities of the image behind distortions caused by the noise in the input. Unlike Variational AutoEncoders (VAEs) \citep{kingma2013auto}, for which the encoder, bottleneck and decoder are defined by their distinct assigned roles through a dual objective function, the components of the UNet denoiser are only defined architecturally, and the optimization is solely driven by a single end-to-end denoising objective. As a result it is not clear where the denoising occurs.

Some insight arises from considering the operation of denoisers designed in the pre-DNN era. Traditionally, image denoisers operate by transforming the noisy image to a latent space in which the true image is concentrated (sparse), and the noise is distributed (dense). Then the noise is suppressed and the image preserved. Finally, the latent representation is transformed back to the image space. This basic description captures the Wiener filter (operating in the frequency domain), and thresholding denoisers which are typically applied within a multi-scale wavelet decomposition \citep{milanfar2012tour}. In all cases, more concentrated signal representations lead to better separation of noise and signal, hence superior denoising. 

\begin{figure}
    \centering
    \includegraphics[width=.8\linewidth]{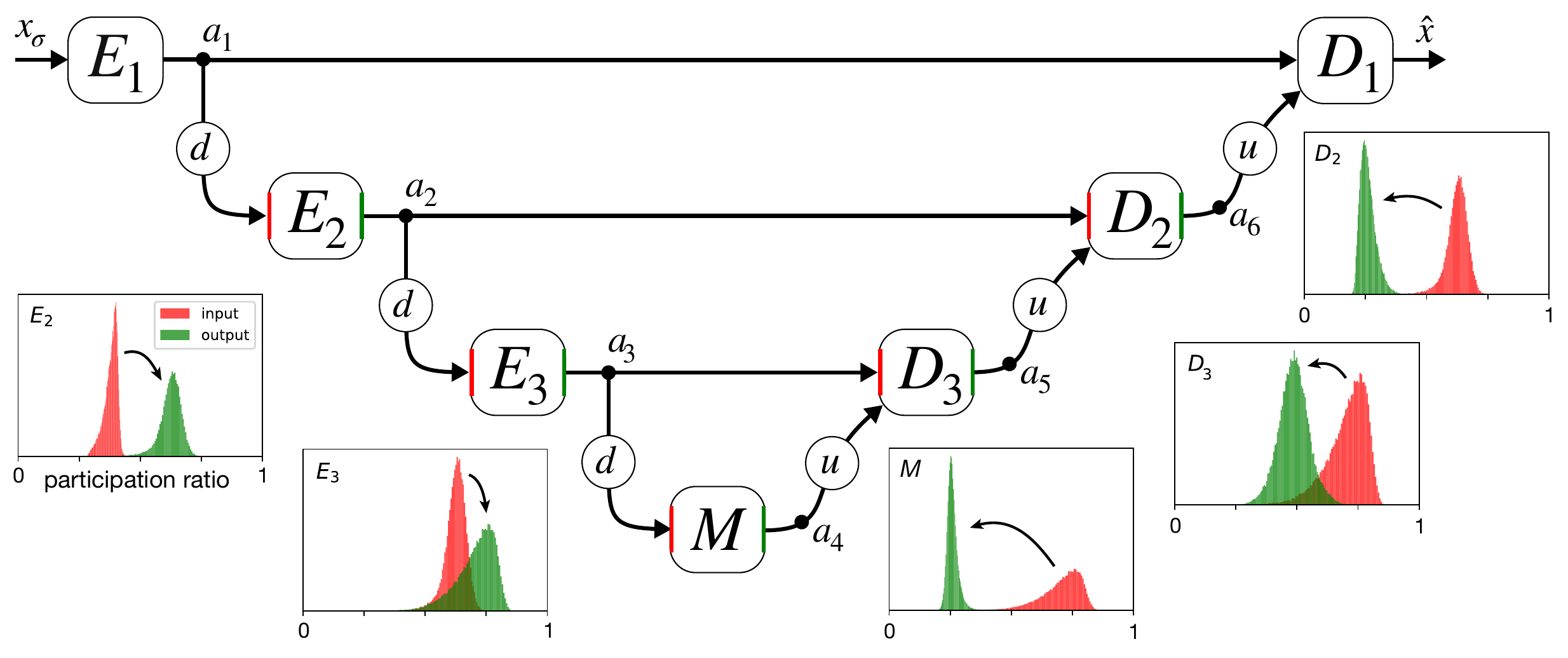}
   \caption{{Channel sparsity of input and output layers of a UNet trained on ImageNet.}
   Histograms show participation ratios (PR), of the spatially averaged input channels (orange) and output channels (green) for individual blocks. The middle block and decoder blocks exhibit increases in sparsity (i.e. reduction in PR). Blocks $\{ E_1, D_1\}$ are not included since they have only one input/output channel, receptively. 
   This is evidence that encoder blocks extract features to isolate noise and signal, and middle block and decoder blocks preserve those channels containing signal while suppressing those containing noise.
   (Notation: encoder blocks $\{E_k\}$, middle block  ($M$), decoder blocks $\{D_k\}$, downsampling ($d$), upsampling ($u$),``skip'' connections.) See \Cref{fig:representation-sparsity-texture} for other models.
   }
    \label{fig:architecture-sparsity}
\end{figure}

In the era of deep learning, DNNs optimized for denoising far outperform traditional solutions. Assuming the same principles apply, we expect to observe an increase in representation sparsity in portions of the network responsible for removing noise. To search for this locus, we measure the sparsity of $\bar{a}$ at the input and output layers of each block of a UNet trained on ImageNet. Sparsity of $\bar{a}$ is quantified by the normalized {\em participation ratio} (PR), the squared ratio of $L_1$ and $L_2$ norms:
\begin{align*}
\text{PR}(\bar{a}) = \frac{\Vert \bar{a}\Vert^2_1}{ d\ \Vert  \bar{a}\Vert^2_2 }.
\end{align*}
where $d$ is the ambient dimensionality of the space, i.e. number of channels in the layer.
$\text{PR} \in (0,1]$, and provides a soft measure of dimensionality: small values reflect low dimensionality (sparsity), and large values indicate high dimensionality (density). 

\Cref{fig:architecture-sparsity} shows distributions of sparsity for input/output layers of  blocks in the UNet denoiser trained on ImageNet64 dataset. 
Both encoding blocks offer a substantial decrease in sparsity of $\bar{a}$, but we observe a stark increase of sparsity in the middle and decoding blocks. This suggests that removal of noise, through suppression of channels whose activations are primarily carrying noise, starts at the middle blocks, and then continues in the decoder blocks. The decoder block $j$, hence, denoises the features present in $a_j$ passed from encoder $j$, conditioned on the denoised features coming from below. This is evidence that the encoder transforms the noisy image to a hierarchical representation in preparation for denoising, by extracting image structures learned from the training set. 
Thus, the representation lends itself to investigation at the output layers of the middle and decoder blocks, where image features are exposed after removal of noise. \Cref{fig:representation-sparsity-texture} shows this phenomenon in three additional UNet models trained on Texture, CelebA, and LSUN-Bedroom datasets.  

%The remainder of this paper is a study of $\bar{a}$, what it represents, and how much of clean underlying image can be recovered from it.  

\begin{figure}
    \centering
    \begin{subfigure}{1\linewidth}    
    \end{subfigure}
     \begin{subfigure}{1\linewidth}
    \includegraphics[width=1\linewidth]{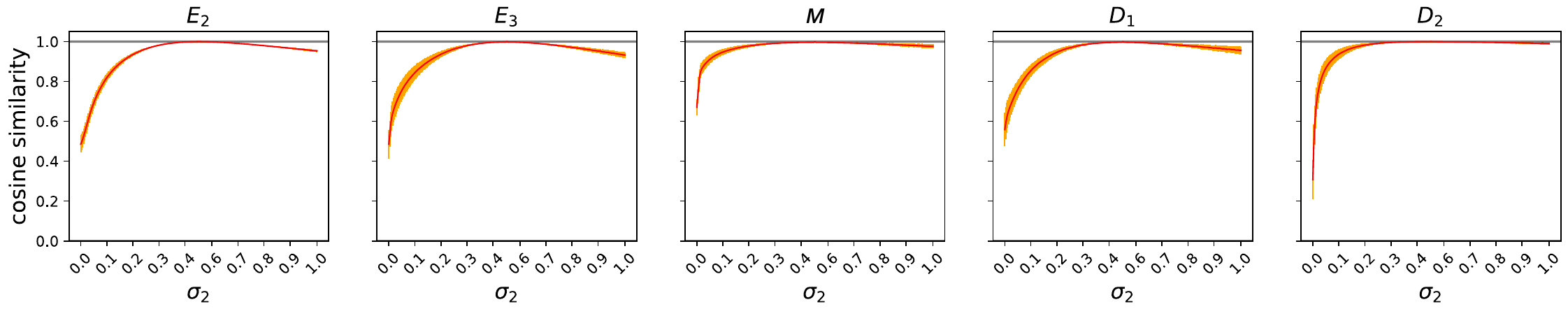}
     \end{subfigure}
    \caption{{Stability of $\bar{a}$ across noise levels, for different network blocks of a model trained on ImageNet64.} Plots show cosine similarity of $\bar{a}(x_{\sigma_1})$ and $\bar{a}(x_{\sigma_2})$, for $\sigma_1 = 0.5$, as a function of $\sigma_2$. $\bar{a}$ is most stable in the middle block (M). Note that $\bar{a}$ collapses as $\sigma$ falls to zero, for which the denoiser should compute the identity function. See \Cref{fig:noise-level-dependency-unet-texture} for other models.}
    \label{fig:noise-level-dependency-unet}
\end{figure}

%%%%%%%%%%%%%%%%%%%%%%%%%%%%%%%%%%%%%%%%%%%%%%%%%%%%%%%%%%%%%%%%%%%%%%%%%
\subsection{Robustness of representation to noise}

To elucidate the relationship between $x$ and $\bar{a}$, we need first to clarify the effects of noise. The vector $\bar{a}$ depends on both the noise amplitude, $\sigma$, and the particular noise realization $x_\sigma$. In the context of diffusion sampling algorithms, the former translates to the evolution of representation with time. Not surprisingly, the variance in $\bar{a}$ grows with noise level. To remove this variability, we take the mean of $\bar{a}$ across noise realizations, $\mathbb{E}_{z}\left[ \bar{a}(x+ \sigma z) \right]$. Figures \ref{fig:noise-level-dependency-unet} and \ref{fig:noise-level-dependency-unet-texture} show the effect of noise level on $\bar{a}$.
Variability due to noise level depends on the block depth, but interestingly, $\bar{a}_j$ is most stable in the middle block: increased noise level mostly increases the amplitude but not the direction of $\bar{a}_j$. The noise resilience of $\bar{a}$ at the output layer of the middle block makes it a good candidate for study. Thus, for the remainder of this paper we focus on the representation in this layer, which we notate as:
\begin{align*}
    \phi(x_\sigma) = \mathbb{E}_{z}\left[ \bar{a}_4(x+\sigma z) \right].
\end{align*}

%%%%%%%%%%%%%%%%%%%%%%%%%%%%%%%%%%%%%%%%%%%%%%%%%%%%%%%%%%%%%%%%%%%%%%%%%
%%%%%%%%%%%%%%%%%%%%%%%%%%%%%%%%%%%%%%%%%%%%%%%%%%%%%%%%%%%%%%%%%%%%%%%%%
% \subsection{Properties of the representation}
% \subsection{Specialized versus common channels}
\subsection{Denoising and channel selectivity}
\label{sec:properties}

\begin{figure}
    \centering
    \includegraphics[width=0.22\linewidth]{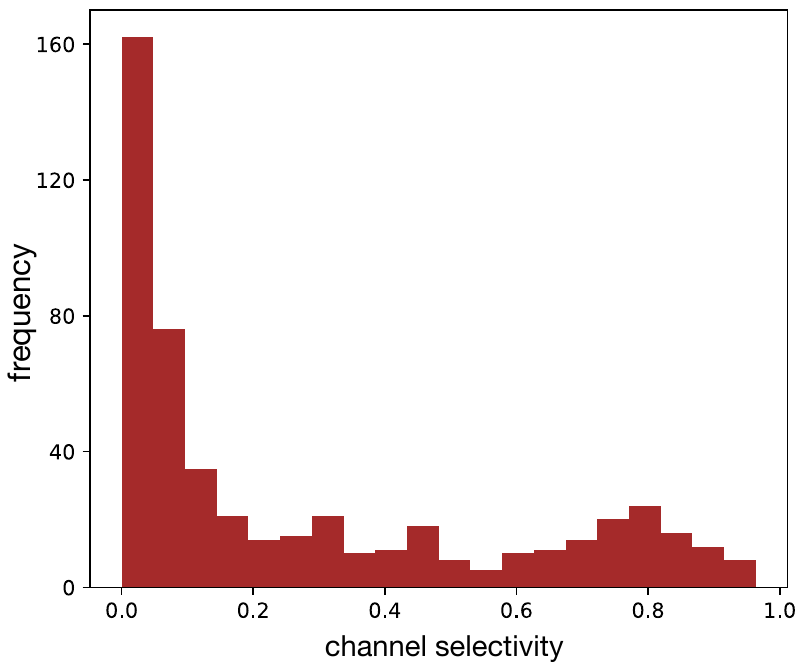}
    \hfil
    \includegraphics[width=0.18\linewidth]{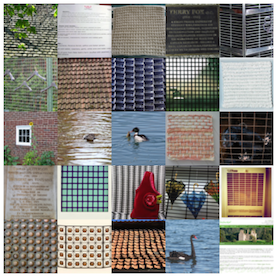}    
    \includegraphics[width=0.18\linewidth]{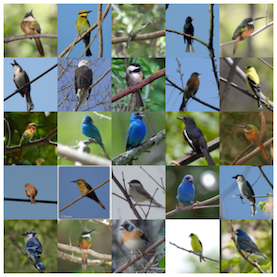}
    \includegraphics[width=0.18\linewidth]{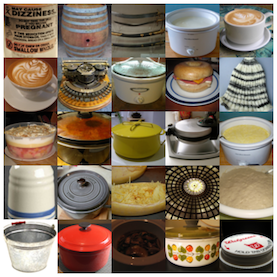}    
    \includegraphics[width=0.18\linewidth]{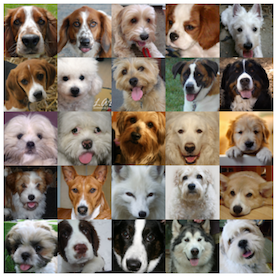}
    \caption{{Channel selectivity.} \textbf{Left:} Participation ratios for each channel over ImageNet. Distribution is bimodal, corresponding to channels that are highly specialized (and infrequently active) on left, and commonly used on right.  
    % for the 512 activations in the middle block computed over ImageNet validation set. Smaller score shows the channel responds to fewer images, hence it is more selective. The distribution is bimodal: channels may be partitioned into a common set (large values) and a specialized set (small values). 
    \textbf{Right:}  
    % Specialized channels respond to highly specific features or patterns, and have negligible responses for most images in the dataset. 
    The panels show the set of images that maximally activate each of four specialized channels, revealing selectivity for rectangular periodic lattices (PR$=0.19$), a bird on a branch ($0.18$), cylindrical objects($0.19$), and dog faces ($0.26$). See \Cref{fig:channel-selectivity-texture} for other models.  
    }
    \label{fig:channel-selectivity}
\end{figure}

% In the previous section, we extracted a sparse representation from the output layer of the middle block. 
In this section, 
% we explore some properties of $\phi$. In particular, 
we examine the components of the representation vector $\phi$, and their relationship to the content of the (clean) image $x$. 
% Then, we study semantic similarities in pixel space in relation to Euclidean distances in the representation space. 
For each channel, $i$, in the output layer of the middle block we quantify its (non-)selectivity using the participation ratio: $\frac{\Vert c(i) \Vert_1^2}{n\ \Vert c(i) \Vert_2^2}$, 
% \begin{align*}
    % \frac{\Vert c(i) \Vert_1^2}{n\ \Vert c(i) \Vert_2^2}
% \end{align*}
where $c(i) = \left (\phi({x_1}_\sigma)[i], ... , \phi({x_n}_\sigma)[i] \right)$ is the concatenation of the $i^{th}$ entry of $\phi$ for all $n$ images in the dataset.
% the ImageNet64 validation set (approximately $50k$ images). 
This provides an estimate of the number of active channels, with smaller values indicating more selectivity. 
% The threshold is set according to activation at zero noise level (Details in Appendix \ref{app:properties representation}). 
Figure \ref{fig:channel-selectivity} shows the distribution of selectivity for all $512$ channels in the middle block of a UNet trained on ImageNet64. The distribution is bimodal. Channels fall roughly into one of the two categories: {\em selective channels} that respond only to a small fraction of images in the set, and {\em non-selective channels} that respond to many images. 
    
\begin{figure}
    \centering
    \includegraphics[width=0.165\linewidth]{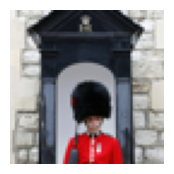}    
    \hspace*{1cm}
    \includegraphics[width=0.195\linewidth]{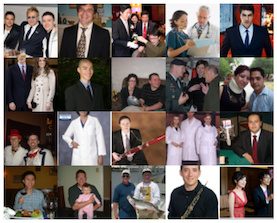}
    \includegraphics[width=0.195\linewidth]{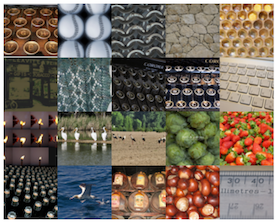}
    \includegraphics[width=0.195\linewidth]{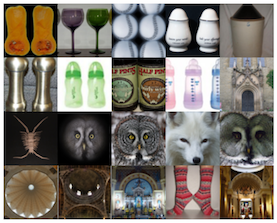}
    \caption{{Specialized channels capture visual attributes and composition of an image.} {\bf Left:} Example image that activates several specialized channels. {\bf Right:} Each panel shows the set of images that maximally activate one of the specialized channels activated by the example image, corresponding to images of people, periodic texture patterns, and images with left-right reflective symmetry. All three elements are present in the example image.}
    \label{fig:compositionality-selective-channels}
\end{figure}

Specialized channels respond to specific features or patterns, and have negligible responses for most images in the dataset \citep{bau2020units}. Figure \ref{fig:channel-selectivity} shows sets of images that maximally excite four example specialized channels. The pattern shared across these images indicates the feature extracted by that channel. 
% within each group clearly share similar patterns.
% The observed sparsity of $\phi$ is a consequence of the existence of specialized channels. 
Since each image only contains a subset of patterns out of all the patterns present in the data distribution, only a handful of specialized channels are activated for each image. Figure \ref{fig:compositionality-selective-channels} shows an example image along with the top three specialized channels it activates. Each channel is selective for a particular property of the image. Over the entire dataset, all channels are used, but different channels are used for different images. 
% \zk{emphasize: non-linear decomposition of features. connect to fourier and wavelet decomp. and their priors }

More generally, channel selectivity predicts some statistical properties of the channels (\Cref{fig:statistics of channels}). The marginal distribution of $\phi$ values in specialized channels is heavy-tailed, but common channels are closer to Gaussian. Specialized channels are spatially sparse while common channels are spatially denser. Additionally, PCA analysis on activation maps of specialized channels shows that they are highly concentrated in a few directions while the common channels are explained with more dimensions. Finally, less selective channels respond to a larger set of frequencies and orientations, and are the most stable w.r.t to noise level. These results imply that the common channels capture shared or common image features, such as brightness, global structure of a scene, etc.

%%The field of image processing has relied on representations optimized for  In the traditional denoising literature, we can find two types of methods: matched filtering or dictionary learning on one hand, which base denoising on pattern detection. These methods can perform well if we have access to the set templates (atoms) that are likely to appear in the image. The problem is that these methods do not scale well, because there is only so many patterns we can include in the dictionary. So they suffer from generalization failures. The second group depends on statistical knowledge of the distribution of all images, like wavelet thresholding methods, GSM or other MRF methods. These are very general, and can be applied to all images, but performance hits a ceiling. It seems like the deep net has learned a mix of the two: the specialized channels are similar to atoms in an adaptive dictionary, while the common channels are more like the axes of a wavelet transform. So the network would not completely fail in generalizing to out of distribution patterns thanks to common channels, but it degrades significantly.  Another conjecture is, increasing increasing the dimensionality of the $\phi$ space will result in more specialized channels. 

%%%%%%%%%%%%%%%%%%%%%%%%%%%%%%%%%%%%%%%%%%%%%%%%%%%%%%%%%%%%%%%%%%%%%%%%%

% \subsection{Distances in representation space}
\subsection{Denoising and emergent semantic similarity}
\label{subsec:Distances in the representations space}
\begin{figure}
    \centering
   \includegraphics[width=.2\linewidth]{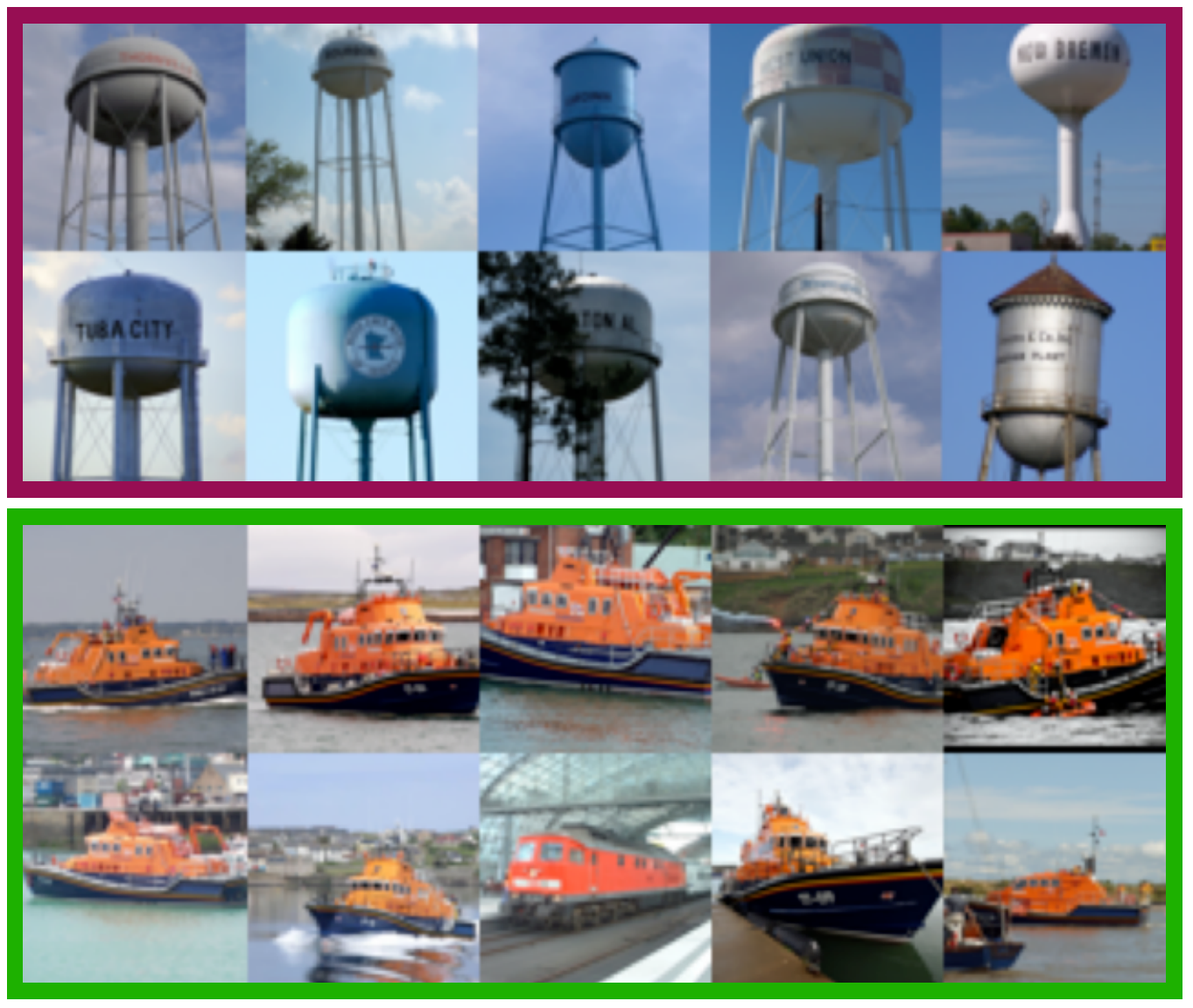}     
   \hfil
    \includegraphics[width=.5\linewidth]{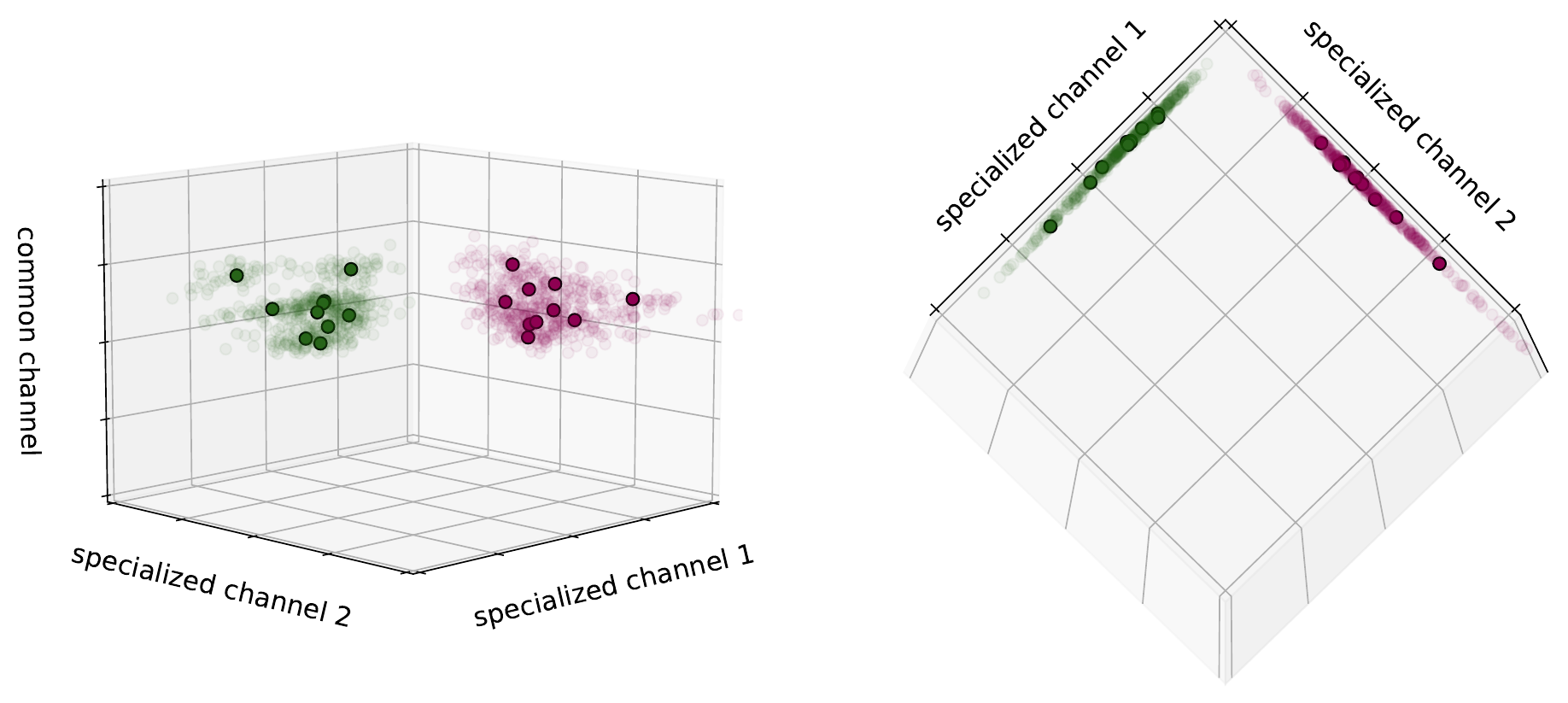}    
    \caption{{Union of subspaces.} {\bf Left:} Two sets of images whose $\phi$'s lie on two subspaces. {\bf Middle/Right:} Three components of the $\phi$ vectors (out of the 512) for these images. The vertical axis corresponds to a common channel, while the other two correspond to specialized channels, each selective for only one image cluster. As a result, the $\phi$ vectors lie on a union of two-dimensional subspaces in the displayed three dimensional ambient space. 
    % Opaque points are computed from individual noise realizations. Discs are centroids of of those.
    }
    \label{fig:union-subspaces}
\end{figure}

An immediate consequence of \emph{sparsity} and \emph{channel selectivity} is that the $\phi$'s lie in a \emph{union of subspaces} in $\mathbb{R}^d$, whose dimensionality lies approximately in the range $[0.2d,0.3d]$ (from Figure \ref{fig:architecture-sparsity}). Each subspace corresponds to a combination of features that are likely to occur simultaneously (Figure \ref{fig:union-subspaces}). What do images whose $\phi$'s lie on the same subspace have in common?
To examine this, we gathered images from the dataset with highest cosine similarity to a given target image in the representation space. Figures \ref{fig:union-subspaces} and \Cref{fig:distances-pixel-latent1} show two sets of those images. Images that are closest in the $\phi$ are semantically similar, in stark contrast to images that are closest in the pixel space. 

% How many of these subspaces are used? On one extreme, if the zero entries of $\phi$  coincide for all images $x$, this number could be as small as one. On the other extreme, it is bounded above by the number of possible  $k$-dimensional subspaces, approximately $\sum_{k=100}^{150}\binom{500}{k}$, which far exceeds the size of any feasible dataset. The assumption behind this extreme case is that features corresponding to entries of $\phi$ are independent from one another, so that all combinations of features are equally likely. We know intuitively that this is not the case. Only certain combinations are likely and frequent in the training data distribution. This implies that collections of $\phi$'s should lie in common subspaces (Figure \ref{fig:union-subspaces}).

% We can associate these samples to the posterior distribution induced by the conditioner image, $p(x|\phi_(x^c))$. From a Bayesian perspective, and in the context of diffusion, the gradient of the prior and likelihood can be employed at all noise levels to sample from the posterior  
% \begin{align*}
% \nabla_{x_{\sigma}} \log p_{\sigma}(x_{\sigma}|\phi_{x^c_\sigma} ) = \nabla_{x_{\sigma}}  \log p_\sigma({x_{\sigma}} ) + \nabla_{x_{\sigma}}  \log p_\sigma(\phi_{x^c_\sigma}|{x_{\sigma} } )
% \end{align*}
% The first term is approximated by a score network, $s(x_\sigma)$. Computing the second term is less straightforward, because the likelihood of $x_\sigma$ given $\phi_(x^c)$ is normally quite complex and cannot be analytically defined, but a Gaussian approximation \sm{of covariance Identity, which is a very strong hypothesis that we know not to be aproximatively valid} leads to 
% \begin{align*}
% \nabla_{x_{\sigma}} \log p_{\sigma}(x_{\sigma}|\phi_{x^c_\sigma} )                                 
% & \approx s({x_{\sigma}} ) + (\phi_{x^c_\sigma} - \phi({x_{\sigma}} ))^T \nabla_{x_\sigma}\phi({x_{\sigma}} )  
% \end{align*}
% The rank of $\nabla_{x_\sigma}\phi({x_{\sigma}} $  determines the dimensionality of the subspace and the top right singular vectors determines the axes of the subspace. This is a matching step with more strength in the direction of larger singular values. These are directions that determine 
% \zk{not finished}
%%%%%%%%%%%%%%%%%%%%%%%%%%%%%%%%%%%%%%%%%%%%%%%%%%%%%%%%%%%%%%%%%%%%%%%%%%%%%%%%%%%%%%

% \subsection{Emergent clusters}
\begin{figure}[t]
    \centering
    \hfil Emergent clusters
    \hfil\hfil
    Human-labeled object classes
    \includegraphics[width=.9\linewidth]{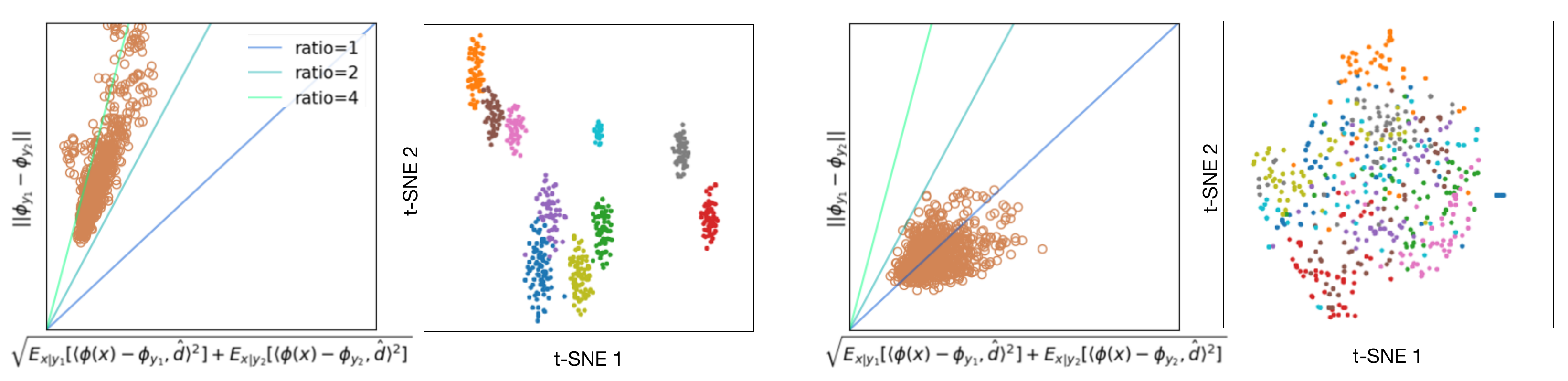}
    % \vspace{-.2cm}
    \foreach \i in {10, 15,18,22,40,24,49} {
        \includegraphics[width=1\textwidth]{figures/mixture_model/embedding/c-\i.png}
    }    
    \caption{Representation vectors, $\phi$, are strongly concentrated within emergent clusters, less so with respect to object classes.
    \textbf{Top Left:} Euclidean distance between centroids of two clusters plotted against square root of their average variance along the line connecting the centroids, for 100 randomly selected cluster pairs. For all pairs, centroids are more than two standard deviations apart, and thus well-separated. t-SNE visualization of 7 randomly selected clusters. \textbf{Top Right:}  Analogous plots for pairs of human-labeled classes, which are substantially more overlapping. \textbf{Bottom:} Each row contains images randomly selected from a cluster.
    Images within a cluster are visually similar and share global organization and semantic patterns, but they not necessarily from the same class. For example, dogs are clustered based on their pose, but not their breed. 
    % Thus, from a learned unconditional denoiser, a new notion of similarity emerges. 
    More examples in Appendix \ref{app:properties representation}. }
    \label{fig:dprime-tSNE}
    % \vspace{-.5cm}
\end{figure}
The fact that pairwise Euclidean distances in the $\phi$ space are meaningful implies that similar images cluster together within the subspace. To test this, we applied a K-Means clustering algorithm \citep{lloyd1982least} to the $\phi$ vectors computed from the ImageNet dataset (See Appendix \ref{app:properties representation}). Figure \ref{fig:dprime-tSNE} shows that  within clusters the $\phi$'s are well concentrated around centroids, and the centroids are well separated: centroids of adjacent clusters are at least 2 standard deviations apart. The $\phi$'s of human labeled classes, by comparison, are less well separated. As a result, we expect images from different classes to appear within the same cluster.
Examples are shown in Figure \ref{fig:dprime-tSNE}. 
% (see Appendix \ref{app:properties representation} for more). 

Images within same clusters are semantically similar: they clearly share location-specific global scene structure, as well as location-nonspecific detailed elements (e.g., objects and their constituent parts, texture). However, in many cases, the similarity is not the object identity: a variety of circular objects appear in one cluster, dogs on a grass background in another, and a variety of bugs on a flat background in another. This reveals that the type of patterns that are useful for computing the score are not object identities.
What connects images within the same cluster is "the gist of the scene" \citep{potter1969recognition,oliva2001modeling,oliva2005gist,oliva2006building,sanocki2023novel}, which is only partially consistent with human labeled class memberships. Remarkably, these \emph{emergent clusters} arise solely from learning the score, by minimizing a denoising objective. 
The network is trained without labels, augmentations, or regularization to induce any specific type of similarity or grouping (e.g membership in the same class). 
Therefore proximity in the representation space defines a \emph{fully unsupervised} form of similarity in the image space. 
% In the unconditional model, there is no external source of information to dictate proximity in the latent space. %This all arises unsupervised from learning the score. 

%%%%%%%%%%%%%%%%%%%%%%%%%%%%%%%%%%%%%%%%%%%%%%%%%%%%%%%%%%%%%%%%%%%%%%%%%
%%%%%%%%%%%%%%%%%%%%%%%%%%%%%%%%%%%%%%%%%%%%%%%%%%%%%%%%%%%%%%%%%%%%%%%%%
\section{Stochastic image reconstruction by conditional sampling}
In order to define the representation more precisely, we propose an algorithm to ``decode'' $\phi$. As with any other representation, decoding $\phi$ is essential to exposing what it represents. Computing $\phi$ involves taking spatial average of the activations.
% , that renders the representation translation invariance. 
Hence, $\phi$ represents not only the image it is computed from, but a whole set of images which have different activations with the same spatial average. In other words, $\phi$ is a many to one function. The goal is to sample from $p(x)$
% the density embedded within the denoiser (or equivalently the score network) 
using a diffusion algorithm, while requiring that all samples have the same $\phi(x^c)$ computed from a target image, $x^c$.
%
% This is a conditional sampling solutions for a non-linear inverse problem. The forward model, however, is not an auxiliary network or external measurement function, but the Encoding half of the same score network. 
Since the algorithm samples images that are consistent with the network's own representation of a target image, it is a \emph{stochastic} reconstruction from the representation. 

The algorithm is obtained by augmenting a reverse diffusion algorithm with a soft projection onto the non-convex set in pixel space defined by $\phi$. 
To achieve this, the score step is alternated with an iterative projection step, where the sample is modified until its representation matches $\phi(x^c)$. This matching step is analogous to methods used to generate texture images from their measured statistics  \citep{portilla2000parametric}. The matching is implemented by minimizing the Euclidean distance between the sample and target representations, $\|\phi(x) - \phi(x^c) \|^2$, which is achieved by back-propagating the gradient of the loss through the first half of the network (i.e. $\bar{a}_4(x_\sigma)$). The algorithm, hence, consists of two alternating steps described in Algorithms \ref{alg:matching} and \ref{alg:sampling}: At every time point in the synthesis, first the $\phi$ of the sample is matched to the target's, via back propagating a gradient in the score network. Then, the score-directed step pushes the sample closer to $p(x)$ by removing noise via using the decoder of the UNet.
%%%%%%%%%%%%%%%%%%%%%%%%%%%%%%%%%%%%%%%%%%%%%%%%%%%%%%%%%%%%%%%%%%%%%%%%%%%%%%%%%%%%%%%%%%%
\noindent
\begingroup
\begin{algorithm}
\caption{Stochastic reconstruction from the representation}
\label{alg:sampling}
\begin{algorithmic} \small
 \Require score network $s$, conditioner image $x^c$, sigma schedule $(\sigma_T, ...,\sigma_0)$, distribution mean $m$
 \State Draw $x_{T} \sim \mathcal{N}(m, \sigma_T^2\mathrm{Id})$ \Comment Initialize sample
 \For{$t \in (T,..., 1)$}  \Comment{Alternate between matching and score steps}
    \State Draw $z \sim \mathcal{N}(0,\mathrm{Id})$ 
    \State $x_t^c = x_c + \sigma_t z$
    \State $x_t \gets $ Guidance step $(x_t, x_t^c)$   \Comment{Update $x_t$ to match representations using Algorithm \ref{alg:matching}}
    \State  \text{Draw} $ z \sim \mathcal{N}( 0,I)$\;               
   \State $x_t \gets x_t  +  s(x_t) +  \sigma_{t-1} z$
   \Comment Update $x_t$ using the score  
\EndFor
    \State $x \gets x_{_1} +  s(x_{1})  $
 \State {\bfseries return} $ x$
\end{algorithmic}
\end{algorithm}
%%%%%%%%%%%%%%%%%%%%%%%%%%%%%%%%%%%%%%%%%%%%%%%%%%%%%%%%%%%%%%%%%%%%%%%%%%%%%%%%%%%%%%%%%%%%%
\begin{algorithm}
\caption{Guidance step}
\label{alg:matching}
\begin{algorithmic} \small
\Require  score network $s$ to compute $\phi(x)$, sample $x_t$, conditioner image  $x^c_t$, learning rate of optimizer $\eta$
\vspace{.1 cm}    
\State $\phi_{x^c_t} = \phi(x^c_t) \hspace{.2cm} \text{and}\hspace{.2cm} \phi_{x_t} = \phi(x_t) $ \Comment{Compute representations}
\State $ \mathcal{L}(x_t) = \Vert \phi(x_t) - \phi_{x^c_t} \Vert^2$   \Comment Compute distance between representations
\State \While{not converged} \Comment{Minimize distance by backpropagating gradient through $e$}         
      % \State \hspace{.5cm} Compute \( \frac{\partial \phi}{\partial x} =\nabla_y\phi({x_{\sigma}} )\)  \Comment{Jacobian of \( e \) w.r.t. \( x \)}
      \State \( \nabla_{x_t} \mathcal{L}(x_t) = (\phi_{x^c_t} - \phi(x_{t}) )^T \nabla_{x_t}\phi(x_{t}) \) \Comment{Compute gradient}
      \State $ x_t \gets x_t - \eta \nabla_{x_t} \mathcal{L}(x_t)$ \Comment{Update}
\EndWhile
\State {\bfseries return} $ x_t $
\end{algorithmic}
\end{algorithm}
\endgroup

%%%%%%%%%%%%%%%%%%%%%%%%%%%%%%%%%%%%%%%%%%%%%%%%%%%%%%%%%%%%%%%%%%%%%%%%%%%%%%%%%%%%%%%%%%%%%

\begin{figure}
    \centering
    \includegraphics[width=0.62\linewidth]{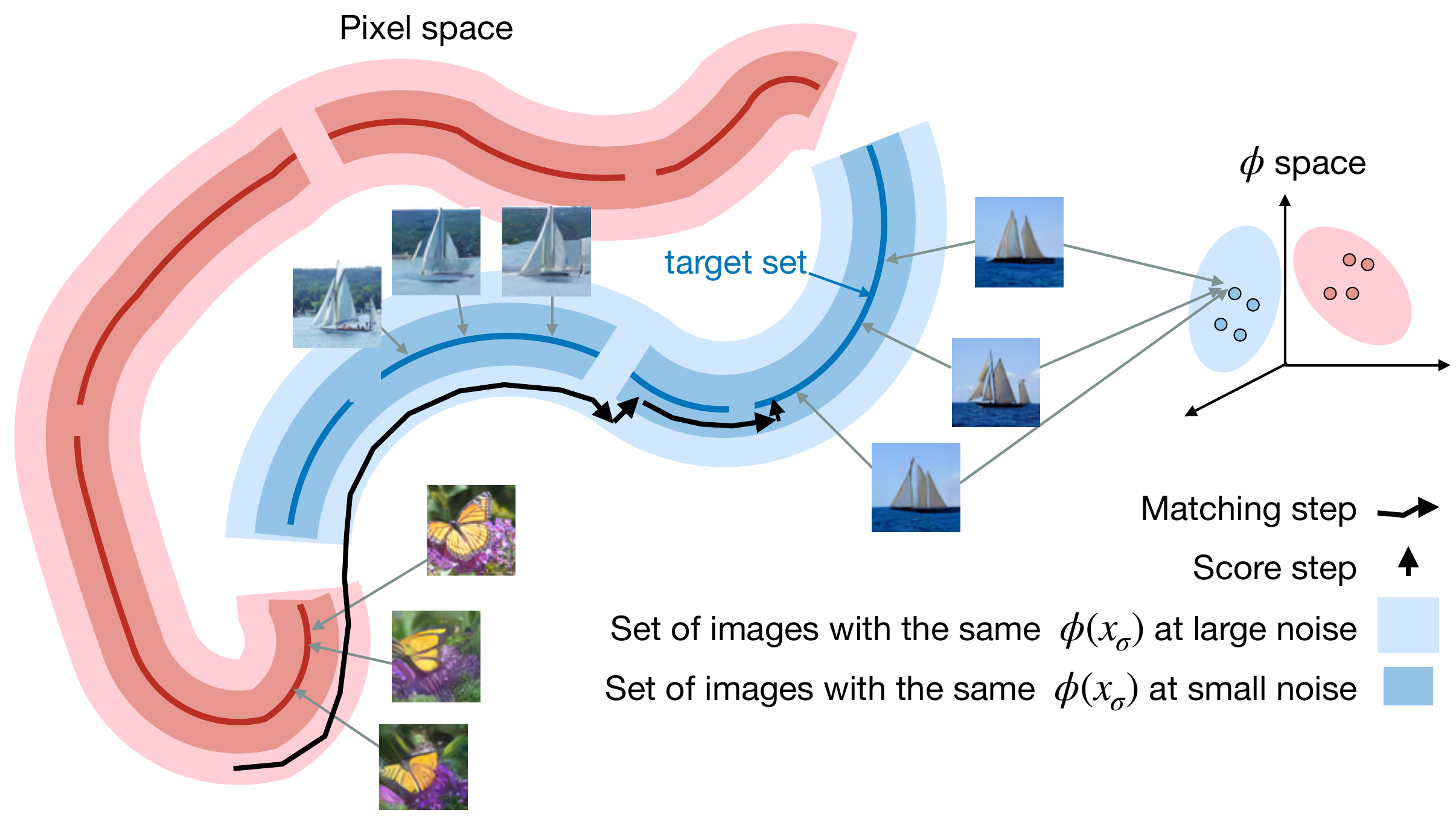}
    \hfill
    \includegraphics[width=0.32\linewidth]{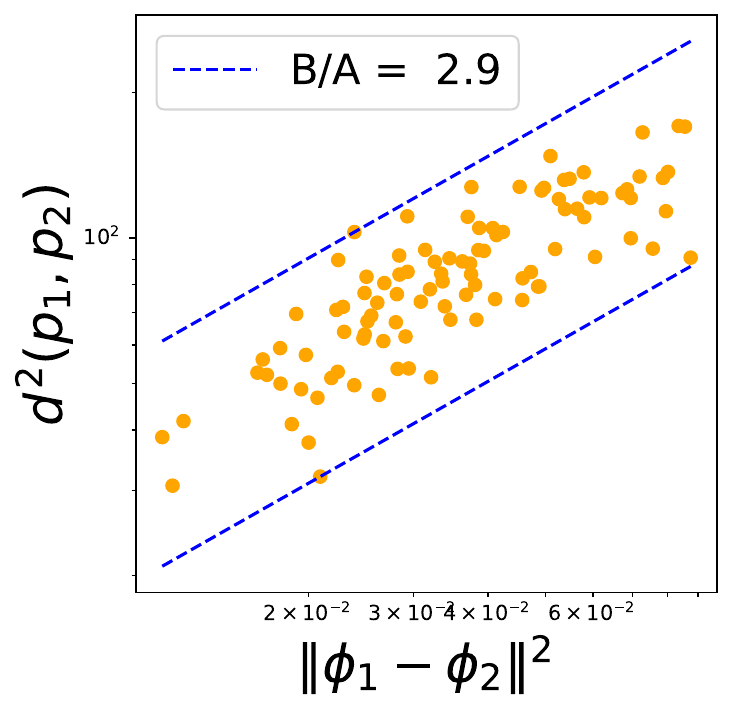}      
    \caption {\textbf{Left:} Schematic of alternating matching and score steps.  \textbf{Right} The distance between random pairs of $\phi$ vectors is strongly related to the distance between the pair of conditional densities they induce.}
    \label{fig:schematic-KL}
\end{figure}

This procedure can be described as a "self-guided" sampling algorithm 
% \citep{dhariwal2021diffusion}
, where the synthesis is guided by the network's own representation. Matching the $\phi$'s changes the trajectory by forcing the sample to be within the set of images whose $\phi$ vectors are equal to the conditioner's. 
% The effect of matching on $x_\sigma$ in one step is visualized in Figures \ref{fig:effect-of-guaidance-digram} and \ref{fig:effect-of-guaidance-exmaples}. 
The matching step does not change the noise level (Figure \ref{fig:noise-change-matching}), so it is in a sense "orthogonal" to the score step. This means that matching iterations guide the sample in the domain of images that are high probability according to $p_\sigma(x_\sigma)$ to reach the set of images defined by $\phi(x_\sigma^c)$. A conceptual diagram of the algorithm is shown in \Cref{fig:schematic-KL}.

\begin{figure}
    \centering
    \includegraphics[width=0.19\linewidth]{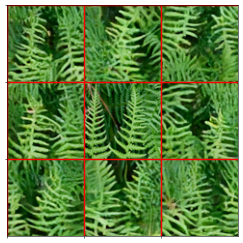}
    \includegraphics[width=0.19\linewidth]{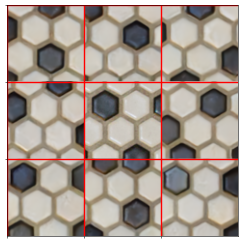}
       \includegraphics[width=0.19\linewidth]{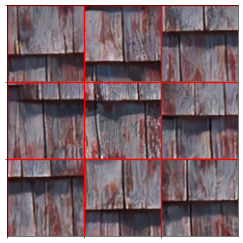}
    \includegraphics[width=0.19\linewidth]{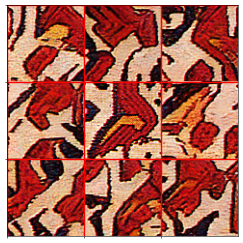}   
    \includegraphics[width=0.19\linewidth]{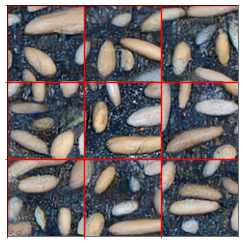}   
    \includegraphics[width=0.19\linewidth]{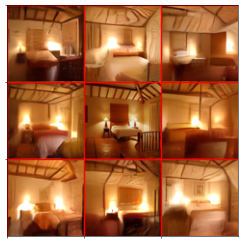}
    \includegraphics[width=0.19\linewidth]{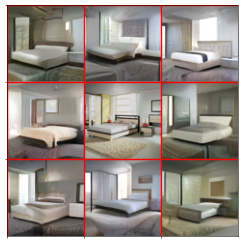}       
    \includegraphics[width=0.19\linewidth]{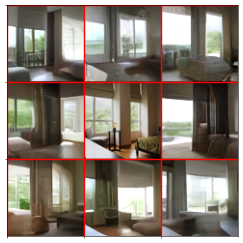}     
    \includegraphics[width=0.19\linewidth]{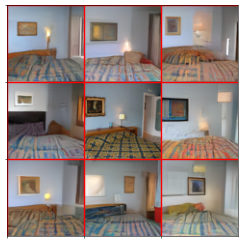}    
    \includegraphics[width=0.19\linewidth]{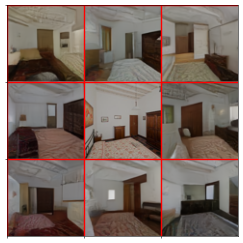}   
    \includegraphics[width=0.19\linewidth]{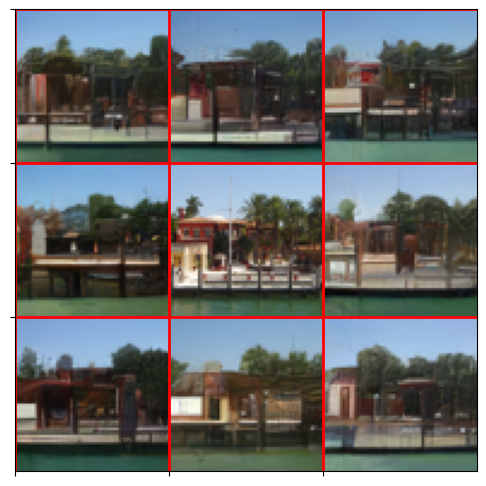}    
    \includegraphics[width=0.19\linewidth]{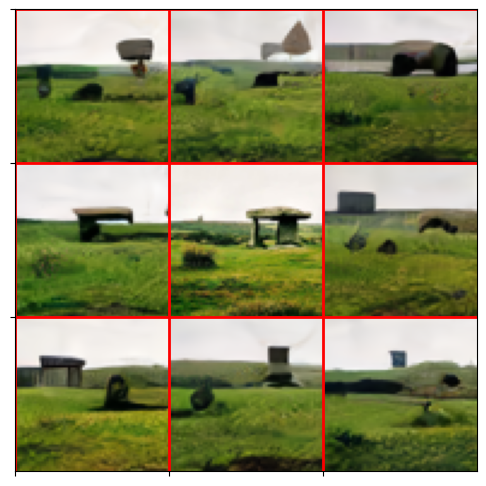}  
    \includegraphics[width=0.19\linewidth]{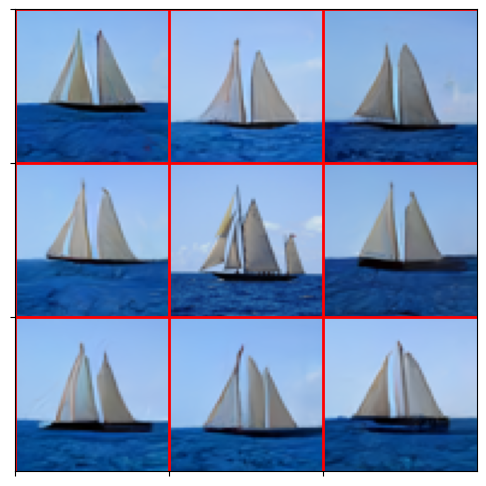}
    \includegraphics[width=0.19\linewidth]{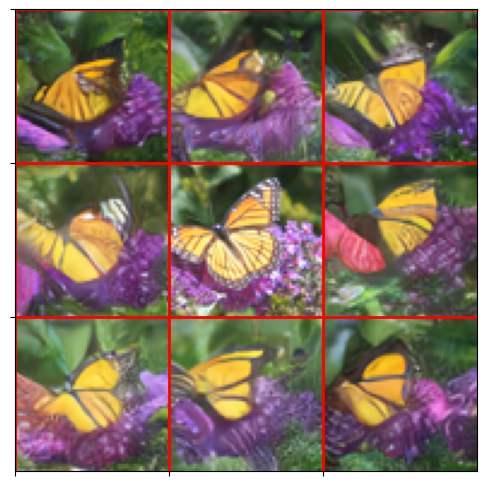}
    \includegraphics[width=0.19\linewidth]{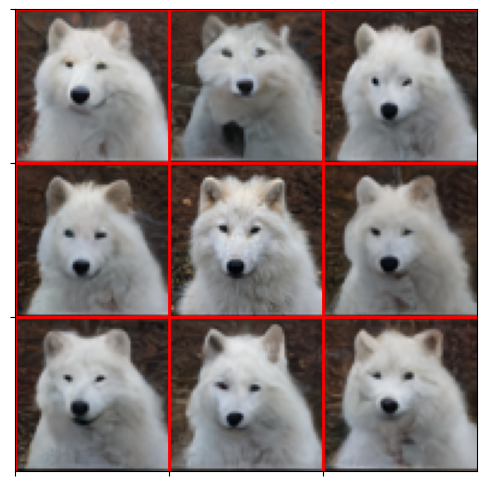}
    \includegraphics[width=0.19\linewidth]{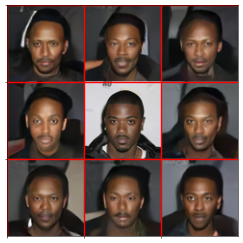}
    \includegraphics[width=0.19\linewidth]{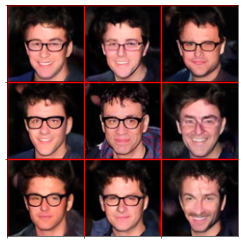}       
    \includegraphics[width=0.19\linewidth]{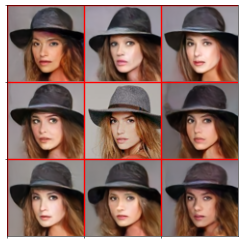}     
    \includegraphics[width=0.19\linewidth]{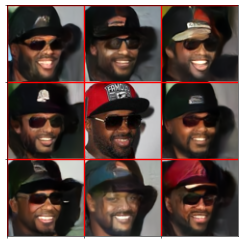}    
    \includegraphics[width=0.19\linewidth]{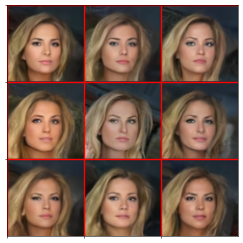}    
    \caption{Samples from the stochastic reconstruction algorithm \ref{alg:sampling}, using models trained \emph{unconditionally} on Texture, LSUN-Bedrooms, ImageNet64, and CelebA datasets (from top to bottom row). Image at center of each panel is an original image from which the target $\phi$ is computed. The surrounding eight images are samples, conditioned on that $\phi$. Semantic similarities between samples within each panel reveal image structures captured by $\phi$. 
    % The diversity of samples varies for different target images, demonstrating that the sets defined by different $\phi$'s vary in size. 
    See Appendix \ref{app:sampling results} for more examples and also samples from the same models without conditioning.}
    \label{fig:samples-imagenet}
\end{figure}

Figure \ref{fig:samples-imagenet} shows samples generated by the algorithm. In each panel, the representation is obtained from a real image. Then the stochastic reconstruction algorithm is used to generate $8$ images with the same $\phi$. The conditionally generated samples are not identical, but all are visually similar to the conditioner image $x^c$. Importantly, they share location-specific global structure as well as location-nonspecific details. Thus, the visual commonalities and diversities in the samples reveal what is and is not captured by the spatial averages of the feature vector in the middle block. 

Each $\phi$ characterizes a conditional density, ${p}(x|\phi)$, from which Algorithm \ref{alg:sampling} draws samples. 
% Distances between these conditional densities should be related to distances between $\phi$'s. 
We show that Euclidean distances between $\phi$'s are correlated with distances between conditional densities they induce. This is captured by a Euclidean embedding property, which ensures that the separation of $\phi$'s is related to a distance between the probability distributions ${p}(x|\phi)$, and hence that there exists $0 < A \leq B$ with $B/A$ not too large, such that
\begin{equation}
      \label{Euclidean-Embedd}
    \forall x_1,x_2~~,~~A \| \phi(x_1) -  \phi(x_2) \|^2 \leq d^2({p}_1,{p}_2) \leq B \| \phi(x_1) -  \phi(x_2)\|^2 .
\end{equation}
We establish a distance between two conditional densities as 
\begin{align}
        d^2({p}_1,{p}_2)  =
        \int_0^{\infty} & \Big( \E_{{p}_1 } \left[\|\nabla \log  {p}_1   - \nabla \log {p}_2 \|^2 \right] 
         +
          \E_{{p}_2 } \left[\|\nabla \log  {p}_1   - \nabla \log {p}_2 \|^2 \right ] \Big)\, \sigma\,d\sigma .
\label{eq:density-distance}
\end{align}
where ${p}_1 = {p}_\sigma(x_\sigma|\phi_1)$ and ${p}_2 = {p}_\sigma(x_\sigma|\phi_2)$. This distance is based on the difference in the expected score assigned to $x_\sigma$ by ${p}(x_\sigma|\phi_1)$ versus ${p}(x_\sigma| \phi_2)$,  integrated across all noise levels. It provide a distance by symmetrizing the Kullback-Leibler divergence $\text{KL}(p\|q)$ between two distributions $p$ and $q$ proved in \citep{Sergio2010,lyu2012interpretation}:
\[
\text{KL}{(p \| q)} = \int_{0}^{\infty} \E_{p_\sigma} [\|\nabla_{x_\sigma} \log {p_\sigma}(x_\sigma) - \nabla_{x_\sigma} \log q_\sigma (x_\sigma)\|^2]\, \sigma d\sigma .
\]

The right panel of \Cref{fig:schematic-KL} confirms that, for a set of random pairs of images, Euclidean distances of their representations satisfy the Euclidean embedding inequality (\ref{Euclidean-Embedd}) with $\frac{B}{A}=2.9$. See \cref{fig:effect-of-guaidance-exmaples} for an intuition about what the symmetrized distance measures.
%%%%%%%%%%%%%%%%%%%%%%%%%%%%%%%%%%%%%%%%%%%%%%%%%%%%%%%%%%%%%%%%%%%%%%%%%%%%%%%%%%%%%%
%%%%%%%%%%%%%%%%%%%%%%%%%%%%%%%%%%%%%%%%%%%%%%%%%%%%%%%%%%%%%%%%%%%%%%%%%%%%%%%%%%%%%%
\section{Discussion}
%Before the advent of deep neural nets, learning densities in high dimensions from data was deemed  impossible.
Generative diffusion models have shown incredible success in learning and sampling from image densities. This feat is due to near minimum mean square error of denoising networks used in these algorithms. 
%which provides an accurate estimation of the score. 
%How can this be achieved from limited data? This is thanks to the alignment of inductive biases of the network and the underlying density. In this work, we set out to understand the features that 
We hypothesized that these networks must construct an internal representation of image features that differentiate signal from noise, and we sought to elucidate that representation.
%, through the interplay between the denoising objective, the inductive biases of the UNet architecture, and the distribution of training images. 
%We look into the internal space of the network to detect where and how image structures are captured? 
%
We found that the vector of spatial averages of channels in the middle block of a trained UNet can provide a signature of this representation. 
% Across a dataset of photographic images, 
The components of this vector are sparse 
% -- a relatively small subset of channels are active for each image --  
% Some channels are "common", in that they are used across a large fraction of images.  Others are "specialized", in that they are used on only a small fraction of images. 
% the score can be estimated because the probability distribution is essentially captured by a low-dimensional sparse representation 
and thus the scores of complex image distributions have a low-dimensional structure, which is made explicit in these networks.
%can be studied through this representation. 
% In addition to the sparsity of the representation vectors, 
We also found that the geometry of the representation space is meaningful and nearby images in representation space are semantically similar in pixel space. We showed that unsupervised clustering of the representation vectors yields well-separated groups of images that are semantically related, but only partially aligned with object identity.
% visually similar and share location-specific global structure and location-nonspecific details. 
We developed a stochastic reconstruction algorithm, and showed that Euclidean distances in representation space are correlated with the symmetrized KL divergence of reconstruction distributions.  
%It defines a Euclidean embedding of KL distances between probabilities conditioned by different representation vectors. Additionally 
These results show that a network trained ``bottom-up'', using only a denoising objective and no external conditioning, labeling, augmentation or regularization, can nevertheless learn a rich and accessible representation of image structure.

\textbf{Limitations and future directions.} 
Many open questions remain to be explored. One concerns a deeper understanding of the geometry of the representation space. We observed that the denoiser transforms a union of manifolds into a union of subspaces. What is the distribution of the dimensionalities of these subspaces, and how many of them, out of all possible ones, are spanned by the $\phi$’s? Importantly, can we estimate the joint distribution of $\phi$’s within these subspaces? A more comprehensive understanding of the latent space could be used to generalize Algorithm~\ref{alg:sampling} to other conditional settings—for example, sampling from an emergent cluster given its centroid, or combining features to create new subspaces, thereby enabling out-of-distribution generalization. Another direction is to investigate the role of the encoder and decoder blocks in representing image structure, and how these representations depend on the noise level. This requires analyzing the interaction between layer depth and noise level, a relationship that is complex but essential to characterize. Finally, while our analysis has focused on fully convolutional neural networks, it is important to examine whether similar representations with comparable properties also arise in more modern architectures such as transformers.

\myComment{
Missing: limitations/caveats/future work.
Examples:
- Role of common vs. specialized channels?  We have anecdotal evidence that common capture large-scale spatial structure, and specialized indicate non-localized presence of specific features.
- Subspace geometry: how precise is statement that phi's lie in subspaces? what is distribution within these subspaces? 
- Subspace dimensionality and numerosity: What is distribution of dimensionality of these subspaces, and how many of the subspaces of this dimensionality range are "used" by imageNet? - Mixtures: how do we mix/control phi's?  what is phi associated with a mixture of images (additive, or merged side-by-side)?
- controlled sampling: how do we control phi's during sampling (to get better samples)?  Enforce sparsity?  Matching to a target subspace?  Or better coarse-to-fine conditioning?
- What is role of later decoder blocks? 
- Can sparse channel representation be found within other DNN denoisers (e.g., dnCNN)?  
finally, extension to more recent archiectures like transformers. Can we e
}
%%%%%%%%%%%%%%%%%%%%%%%%%%%%%%%%%%%%%%%%%%%%%%%%%%%%%%%%%%%%%%%%%%%%%%%%%%%%%%%%%%%%%%

% \section*{Reproducibility statement}
% All the results in this paper are fully reproducible: The datasets used are publicly available (expect for the Texture dataset). The link to the code repository on GitHub will be released after the review process to preserve anonymity. 

%%%%%%%%%%%%%%%%%%%%%%%%%%%%%%%%%%%%%%%%%%%%%%%%%%%%%%%%%%%%%%%%%%%%%%%%%%%%%%%%%%%%%%
\bibliographystyle{unsrtnat}
\bibliography{main}

\begin{thebibliography}{36}
\providecommand{\natexlab}[1]{#1}
\providecommand{\url}[1]{\texttt{#1}}
\expandafter\ifx\csname urlstyle\endcsname\relax
  \providecommand{\doi}[1]{doi: #1}\else
  \providecommand{\doi}{doi: \begingroup \urlstyle{rm}\Url}\fi

\bibitem[Sohl-Dickstein et~al.(2015)Sohl-Dickstein, Weiss, Maheswaranathan, and Ganguli]{sohlDickstein15}
J~Sohl-Dickstein, E~Weiss, N~Maheswaranathan, and S~Ganguli.
\newblock Deep unsupervised learning using nonequilibrium thermodynamics.
\newblock In Francis Bach and David Blei, editors, \emph{Proc 32nd Int'l Conf on Machine Learning (ICML)}, volume~37 of \emph{Proceedings of Machine Learning Research}, pages 2256--2265, Lille, France, 07--09 Jul 2015. PMLR.

\bibitem[Song and Ermon(2019)]{song2019generative}
Yang Song and Stefano Ermon.
\newblock Generative modeling by estimating gradients of the data distribution.
\newblock \emph{Advances in neural information processing systems}, 32, 2019.

\bibitem[Ho et~al.(2020)Ho, Jain, and Abbeel]{ho2020denoising}
Jonathan Ho, Ajay Jain, and Pieter Abbeel.
\newblock Denoising diffusion probabilistic models.
\newblock \emph{Advances in neural information processing systems}, 33:\penalty0 6840--6851, 2020.

\bibitem[Croitoru et~al.(2023)Croitoru, Hondru, Ionescu, and Shah]{surveyDiffuionVision}
Florinel-Alin Croitoru, Vlad Hondru, Radu~Tudor Ionescu, and Mubarak Shah.
\newblock Diffusion models in vision: A survey.
\newblock \emph{IEEE Transactions on Pattern Analysis and Machine Intelligence}, 45\penalty0 (9):\penalty0 10850--10869, 2023.
\newblock \doi{10.1109/TPAMI.2023.3261988}.

\bibitem[Ronneberger et~al.(2015)Ronneberger, Fischer, and Brox]{ronneberger2015u}
O~Ronneberger, P~Fischer, and T~Brox.
\newblock U-net: Convolutional networks for biomedical image segmentation.
\newblock In \emph{Int'l Conf Medical Image Computing and Computer-assisted Intervention}, pages 234--241. Springer, 2015.

\bibitem[Brempong et~al.(2022)Brempong, Kornblith, Chen, Parmar, Minderer, and Norouzi]{Brempong_2022_CVPR}
Emmanuel~Asiedu Brempong, Simon Kornblith, Ting Chen, Niki Parmar, Matthias Minderer, and Mohammad Norouzi.
\newblock Denoising pretraining for semantic segmentation.
\newblock In \emph{Proceedings of the IEEE/CVF Conference on Computer Vision and Pattern Recognition (CVPR) Workshops}, pages 4175--4186, June 2022.

\bibitem[Yang and Wang(2023)]{yang2023diffusion}
Xingyi Yang and Xinchao Wang.
\newblock Diffusion model as representation learner.
\newblock In \emph{Proceedings of the IEEE/CVF International Conference on Computer Vision}, pages 18938--18949, 2023.

\bibitem[Xiang et~al.(2023)Xiang, Yang, Huang, and Wang]{xiang2023denoising}
Weilai Xiang, Hongyu Yang, Di~Huang, and Yunhong Wang.
\newblock Denoising diffusion autoencoders are unified self-supervised learners.
\newblock In \emph{Proceedings of the IEEE/CVF International Conference on Computer Vision}, pages 15802--15812, 2023.

\bibitem[Baranchuk et~al.(2021)Baranchuk, Voynov, Rubachev, Khrulkov, and Babenko]{baranchuklabel}
Dmitry Baranchuk, Andrey Voynov, Ivan Rubachev, Valentin Khrulkov, and Artem Babenko.
\newblock Label-efficient semantic segmentation with diffusion models.
\newblock In \emph{International Conference on Learning Representations}, 2021.

\bibitem[Mukhopadhyay et~al.(2023)Mukhopadhyay, Gwilliam, Agarwal, Padmanabhan, Swaminathan, Hegde, Zhou, and Shrivastava]{mukhopadhyay2023diffusion}
Soumik Mukhopadhyay, Matthew Gwilliam, Vatsal Agarwal, Namitha Padmanabhan, Archana Swaminathan, Srinidhi Hegde, Tianyi Zhou, and Abhinav Shrivastava.
\newblock Diffusion models beat gans on image classification.
\newblock \emph{arXiv preprint arXiv:2307.08702}, 2023.

\bibitem[Miyasawa(1961)]{Miyasawa61}
K~Miyasawa.
\newblock An empirical {Bayes} estimator of the mean of a normal population.
\newblock \emph{Bull. Inst. Internat. Statist.}, 38:\penalty0 181--188, 1961.

\bibitem[Robbins(1956)]{Robbins1956Empirical}
H~Robbins.
\newblock An empirical bayes approach to statistics.
\newblock In \emph{Proc Third Berkeley Symposium on Mathematical Statistics and Probability}, volume~I, pages 157--163. University of CA Press, 1956.

\bibitem[Efron(2011)]{efron2011tweedie}
Bradley Efron.
\newblock Tweedie's formula and selection bias.
\newblock \emph{Journal of the American Statistical Association}, 106\penalty0 (496):\penalty0 1602--1614, 2011.

\bibitem[Song et~al.(2020)Song, Sohl-Dickstein, Kingma, Kumar, Ermon, and Poole]{song2020score}
Yang Song, Jascha Sohl-Dickstein, Diederik~P Kingma, Abhishek Kumar, Stefano Ermon, and Ben Poole.
\newblock Score-based generative modeling through stochastic differential equations.
\newblock \emph{arXiv preprint arXiv:2011.13456}, 2020.

\bibitem[Kadkhodaie and Simoncelli(2020)]{kadkhodaie2020solving}
Zahra Kadkhodaie and Eero~P Simoncelli.
\newblock Solving linear inverse problems using the prior implicit in a denoiser.
\newblock \emph{arXiv preprint arXiv:2007.13640}, 2020.

\bibitem[Julesz(1962)]{julesz1962visual}
Bela Julesz.
\newblock Visual pattern discrimination.
\newblock \emph{IRE transactions on Information Theory}, 8\penalty0 (2):\penalty0 84--92, 1962.

\bibitem[Zhu et~al.(1998)Zhu, Wu, and Mumford]{zhu1998filters}
Song~Chun Zhu, Yingnian Wu, and David Mumford.
\newblock Filters, random fields and maximum entropy (frame): Towards a unified theory for texture modeling.
\newblock \emph{International Journal of Computer Vision}, 27:\penalty0 107--126, 1998.

\bibitem[Portilla and Simoncelli(2000)]{portilla2000parametric}
Javier Portilla and Eero~P Simoncelli.
\newblock A parametric texture model based on joint statistics of complex wavelet coefficients.
\newblock \emph{International journal of computer vision}, 40:\penalty0 49--70, 2000.

\bibitem[Victor et~al.(2017)Victor, Conte, and Chubb]{victor2017textures}
Jonathan~D Victor, Mary~M. Conte, and Charles~F. Chubb.
\newblock Textures as probes of visual processing.
\newblock \emph{Annual Reviews of Vision Science}, 3:\penalty0 275–--296, 2017.
\newblock \doi{10.1146/annurev- vision- 102016- 061316}.

\bibitem[Kadkhodaie et~al.(2023)Kadkhodaie, Guth, Mallat, and Simoncelli]{kadkhodaie2023learning}
Zahra Kadkhodaie, Florentin Guth, St{\'e}phane Mallat, and Eero~P Simoncelli.
\newblock Learning multi-scale local conditional probability models of images.
\newblock In \emph{The Eleventh International Conference on Learning Representations}, 2023.

\bibitem[Kamb and Ganguli(2024)]{kamb2024analytic}
Mason Kamb and Surya Ganguli.
\newblock An analytic theory of creativity in convolutional diffusion models.
\newblock \emph{arXiv preprint arXiv:2412.20292}, 2024.

\bibitem[Kingma et~al.(2013)Kingma, Welling, et~al.]{kingma2013auto}
Diederik~P Kingma, Max Welling, et~al.
\newblock Auto-encoding variational bayes, 2013.

\bibitem[Milanfar(2012)]{milanfar2012tour}
Peyman Milanfar.
\newblock A tour of modern image filtering: New insights and methods, both practical and theoretical.
\newblock \emph{IEEE signal processing magazine}, 30\penalty0 (1):\penalty0 106--128, 2012.

\bibitem[Bau et~al.(2020)Bau, Zhu, Strobelt, Lapedriza, Zhou, and Torralba]{bau2020units}
David Bau, Jun-Yan Zhu, Hendrik Strobelt, Agata Lapedriza, Bolei Zhou, and Antonio Torralba.
\newblock Understanding the role of individual units in a deep neural network.
\newblock \emph{Proceedings of the National Academy of Sciences}, 2020.
\newblock ISSN 0027-8424.
\newblock \doi{10.1073/pnas.1907375117}.

\bibitem[Lloyd(1982)]{lloyd1982least}
Stuart Lloyd.
\newblock Least squares quantization in pcm.
\newblock \emph{IEEE transactions on information theory}, 28\penalty0 (2):\penalty0 129--137, 1982.

\bibitem[Potter and Levy(1969)]{potter1969recognition}
Mary~C Potter and Ellen~I Levy.
\newblock Recognition memory for a rapid sequence of pictures.
\newblock \emph{Journal of experimental psychology}, 81\penalty0 (1):\penalty0 10, 1969.

\bibitem[Oliva and Torralba(2001)]{oliva2001modeling}
Aude Oliva and Antonio Torralba.
\newblock Modeling the shape of the scene: A holistic representation of the spatial envelope.
\newblock \emph{International journal of computer vision}, 42:\penalty0 145--175, 2001.

\bibitem[Oliva(2005)]{oliva2005gist}
Aude Oliva.
\newblock Gist of the scene.
\newblock In \emph{Neurobiology of attention}, pages 251--256. Elsevier, 2005.

\bibitem[Oliva and Torralba(2006)]{oliva2006building}
Aude Oliva and Antonio Torralba.
\newblock Building the gist of a scene: The role of global image features in recognition.
\newblock \emph{Progress in brain research}, 155:\penalty0 23--36, 2006.

\bibitem[Sanocki et~al.(2023)Sanocki, Nguyen, Shultz, and Defant]{sanocki2023novel}
T~Sanocki, T~Nguyen, S~Shultz, and J~Defant.
\newblock Novel scene understanding, from gist to elaboration.
\newblock \emph{Visual Cognition}, 31\penalty0 (3):\penalty0 188--215, 2023.

\bibitem[Verdú(2010)]{Sergio2010}
Sergio Verdú.
\newblock Mismatched estimation and relative entropy.
\newblock \emph{IEEE Transactions on Information Theory}, 56\penalty0 (8):\penalty0 3712--3720, 2010.
\newblock \doi{10.1109/TIT.2010.2050800}.

\bibitem[Lyu(2012)]{lyu2012interpretation}
Siwei Lyu.
\newblock Interpretation and generalization of score matching.
\newblock \emph{arXiv preprint arXiv:1205.2629}, 2012.

\bibitem[Deng et~al.(2009)Deng, Dong, Socher, Li, Li, and Fei-Fei]{imagenet_cvpr09}
J.~Deng, W.~Dong, R.~Socher, L.-J. Li, K.~Li, and L.~Fei-Fei.
\newblock {ImageNet: A Large-Scale Hierarchical Image Database}.
\newblock In \emph{CVPR09}, 2009.

\bibitem[Yu et~al.(2015)Yu, Seff, Zhang, Song, Funkhouser, and Xiao]{yu2015lsun}
Fisher Yu, Ari Seff, Yinda Zhang, Shuran Song, Thomas Funkhouser, and Jianxiong Xiao.
\newblock Lsun: Construction of a large-scale image dataset using deep learning with humans in the loop.
\newblock \emph{arXiv preprint arXiv:1506.03365}, 2015.

\bibitem[Liu et~al.(2015)Liu, Luo, Wang, and Tang]{liu2015faceattributes}
Ziwei Liu, Ping Luo, Xiaogang Wang, and Xiaoou Tang.
\newblock Deep learning face attributes in the wild.
\newblock In \emph{Proceedings of International Conference on Computer Vision (ICCV)}, December 2015.

\bibitem[Kadkhodaie and Simoncelli(2021)]{kadkhodaie2021stochastic}
Zahra Kadkhodaie and Eero~P Simoncelli.
\newblock Stochastic solutions for linear inverse problems using the prior implicit in a denoiser.
\newblock \emph{Advances in Neural Information Processing Systems}, 34:\penalty0 13242--13254, 2021.

\end{thebibliography}

%%%%%%%%%%%%%%%%%%%%%%%%%%%%%%%%%%%%%%%%%%%%%%%%%%%%%%%%%%%%%%%%%%%%%%%%%%%%%%%%%%%%%%
%%%%%%%%%%%%%%%%%%%%%%%%%%%%%%%%%%%%%%%%%%%%%%%%%%%%%%%%%%%%%%%%%%%%%%%%%%%%%%%%%%%%%%
\newpage
\appendix

\section{Architecture, Dataset, Training}
\label{app:architetcure,training,dataset}
\subsection{Architecture}
We used the original UNet architecture \citep{ronneberger2015u} with a few modifications. The architecture is made up of 3 encoder blocks, 1 middle block, and 3 decoder blocks. The number of layers in each block is 2, 3, and 6 for the encoder, middle and decoder blocks. Each layer consists of $3\times3$ convolutions with zero boundary handling, Layer normalization and ReLU. The number of channels is 64 in the first encoder block and then grows by a factor of two after each downsampling operation, and decreased by a factor of two after each upsampling operation. This results in {64, 128, 256} channels in the encoder blocks, 512 channels in the middle block, and {256, 128, 64} in the decoding blocks. The total number of parameters is $\sim 13m$. 

The \textbf{receptive field size} at the end of the middle block is $84\times84$ surpassing the spatial size of the input images in the ImagNet dataset. This is required for capturing large structures in the absence of attention blocks \citep{kadkhodaie2023learning}. The depth of decoder blocks is increased in order to increase the expressivity of the denoising operations. This is again related to the size of the receptive field at the end of each decoder block with respect to the representation at the end of the middle block.

\begin{figure}[H]
    \centering
    \includegraphics[width=1\linewidth]{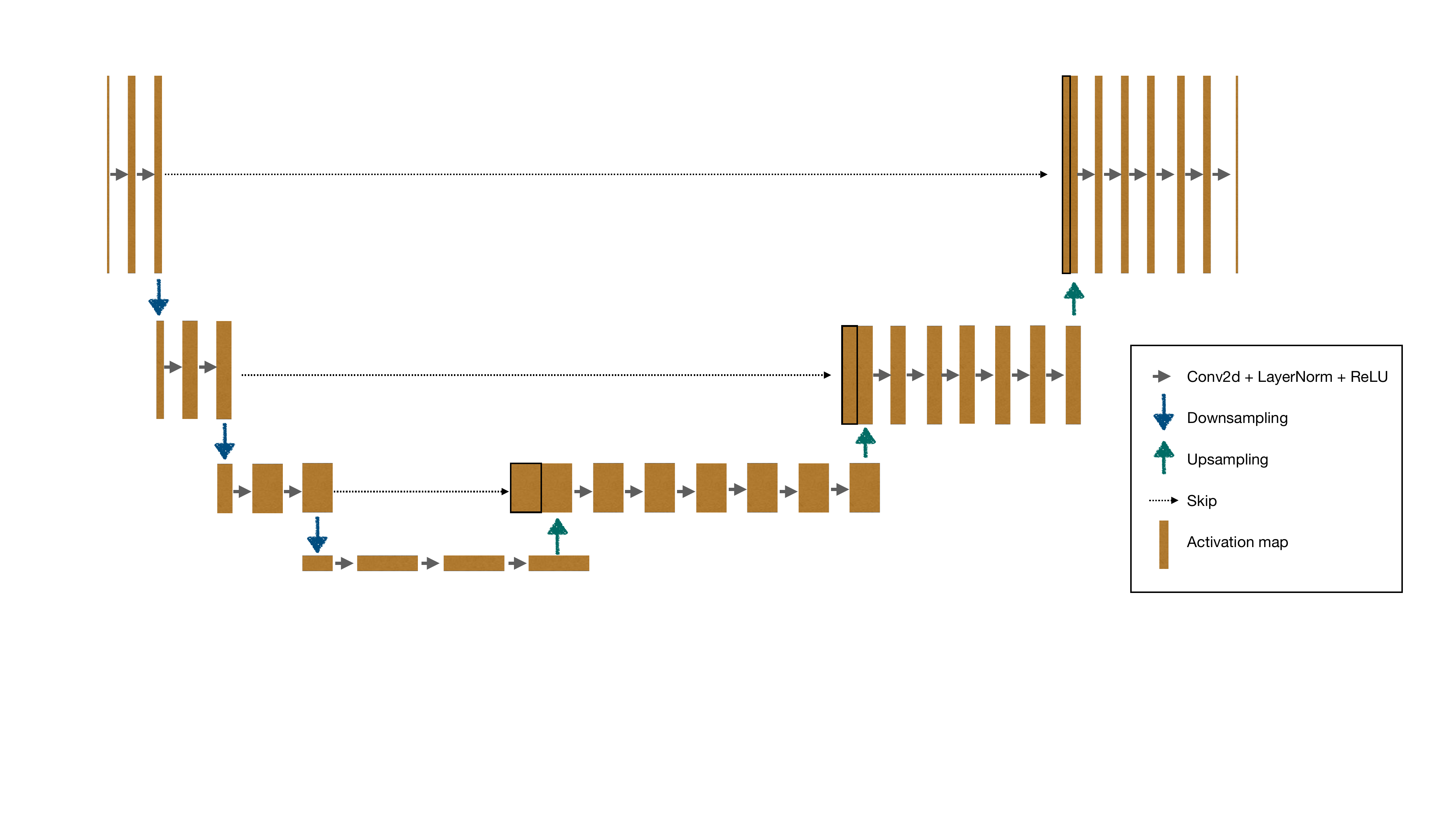}
    \caption{Fully convolutional UNet architecture used in our experiments.}
    \label{fig:unet-architecture}
\end{figure}

\subsection{Datasets} 
We show results from models trained on 4 public datasets: 
\begin{itemize}
    \item ImageNet64: down-sampled version of ImageNet data set \citep{imagenet_cvpr09} , $3\times 64\times64$. The training set consists of $\sim 1.2 m$ images and the validation set consists of $50k$ images, of objects, animals, scenes, etc. We did not use the class labels for training. 
    \item LSUN-Bedroom dataset \citep{yu2015lsun}: down-sampled to $3\times 80 \times 80$ images. We trained the model on a subset of images randomly selected. Training set size $\sim 300,000$ images. 
    \item CelebA dataset \citep{liu2015faceattributes}: down-sampled to $3\times 80 \times 80$ images. Training set size $\sim 200,000$ images. 
    \item Texture dataset (collected by the authors, not published), cropped to $3\times 80 \times 80$ images. Training set size $\sim 230,000$ images.
\end{itemize}

\subsection{Training} 
The network was trained to minimize mean squared loss using Adam optimizer, with an initial learning rate of 0.001 with a decay of a factor of 2 every 100 epochs. Total number of epochs was set to 1000. The size of each batch was $1024$. 

The standard deviation of noise was drawn randomly for each image from a $\frac{1}{\sqrt{\sigma}}$ distribution to emphasize the small noise levels, since we have observed that the it takes more epochs for the denoising MSE to plateau for small noise levels under uniform $\sigma$ distribution.

%%%%%%%%%%%%%%%%%%%%%%%%%%%%%%%%%%%%%%%%%%%%%%%%%%%%%%%%%%%%%%%%%%%%%%%%%%
\section{Properties of representation}
\label{app:properties representation}

\subsection{Representation sparsity}
\begin{figure}[H]
    \centering
    \begin{subfigure}{1\linewidth}    
    \end{subfigure}
     \begin{subfigure}{1\linewidth}
    \includegraphics[width=1\linewidth]{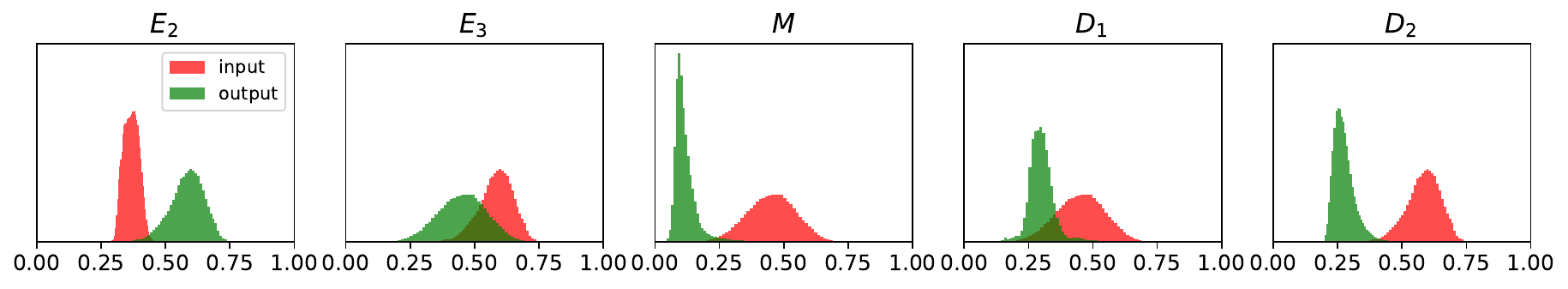}
    \includegraphics[width=1\linewidth]{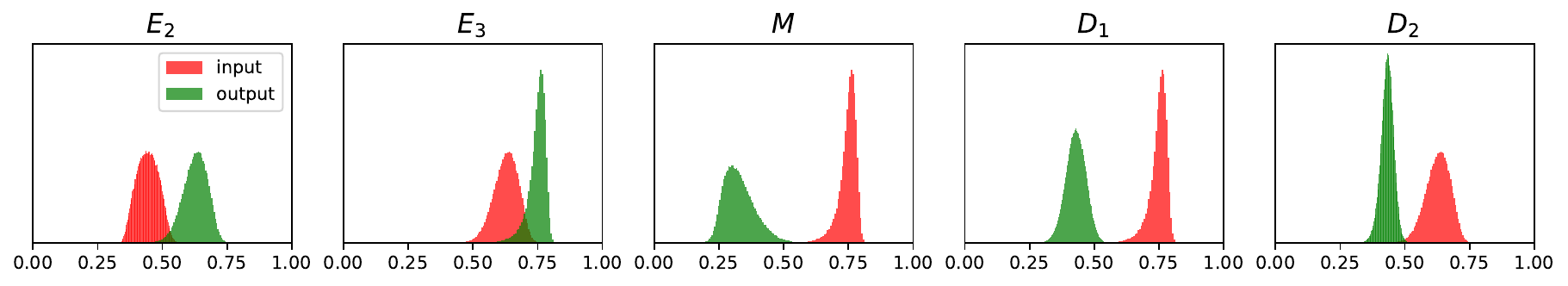}
    \includegraphics[width=1\linewidth]{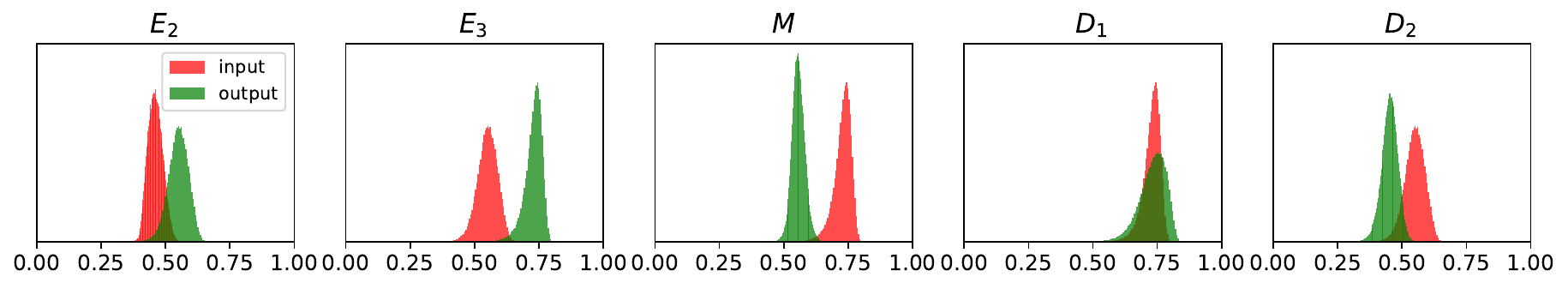}    
     \end{subfigure}
    \caption{Change in sparsity for representations in models trained on \textbf{Texture}, \textbf{LSUN-Bedroom}, and \textbf{CelebA} datasets, from top to bottom row. Channel sparsity increases in the Middle block across dataset. See caption \Cref{fig:architecture-sparsity} for detailed description. }
    \label{fig:representation-sparsity-texture}
\end{figure}

\begin{figure}[H]
    \centering
    \includegraphics[width=.9\linewidth]{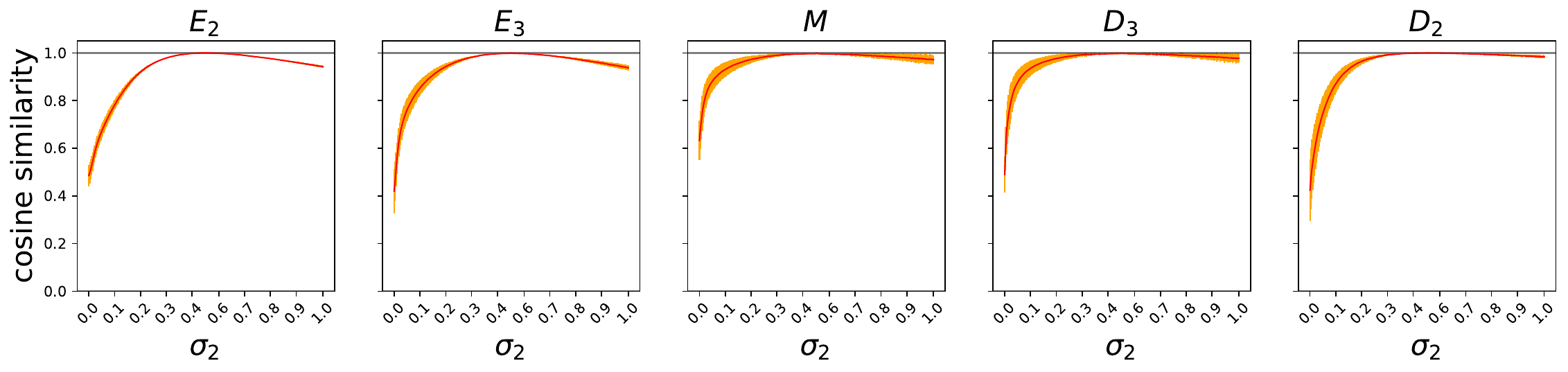}
    \includegraphics[width=.9\linewidth]{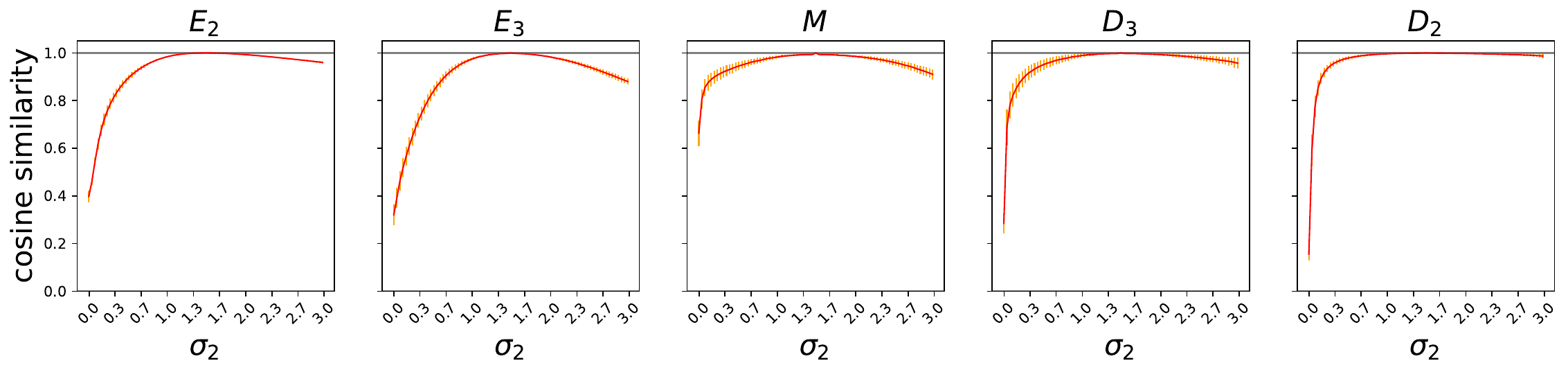}
    \includegraphics[width=.9\linewidth]{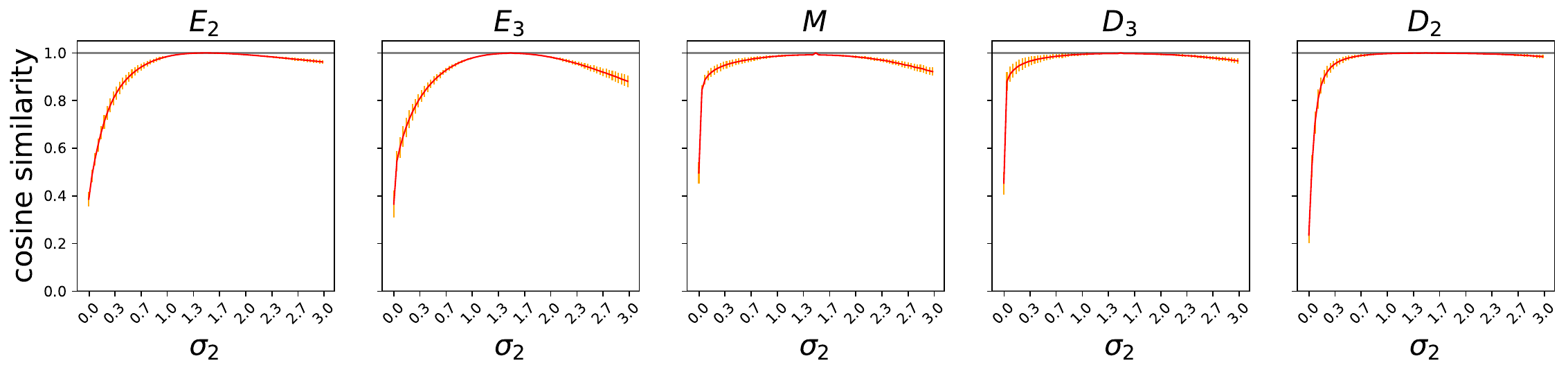}    
    \caption{{Stability of $\bar{a}$ across noise levels, for different network blocks trained on \textbf{Texture}, \textbf{LSUN-Bedroom}, and \textbf{CelebA} dataset.} See caption of \Cref{fig:noise-level-dependency-unet} for detailed description. 
    }
    \label{fig:noise-level-dependency-unet-texture}
\end{figure}

\subsection{Channel selectivity}
% \paragraph{Threshold.} To set the threshold, when computing the selectivity of channels, \Cref{fig:channel-selectivity}, we compute the $\phi$'s at zero noise level for all dataset. In the ideal case, $\phi$ should be zero, but in practice some almost 1/3 of the channels are near-zero. This small amplitude can serve as a channel-wise threshold: For each channel, we set the threshold to be equal to maximum average activation for the channel at $\sigma=0$. 

\begin{figure}[H]
    \centering
    \includegraphics[width=0.22\linewidth]{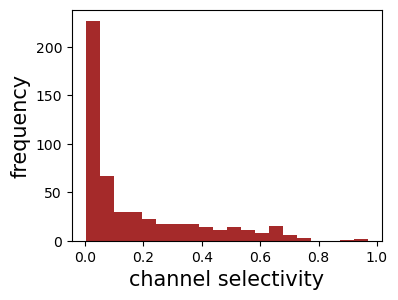}
    \hfil
    \includegraphics[width=0.18\linewidth]{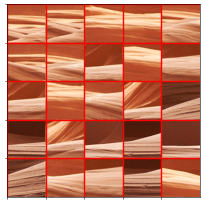}    
    \includegraphics[width=0.18\linewidth]{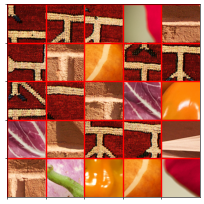}
    \includegraphics[width=0.18\linewidth]{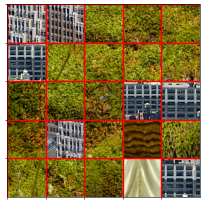}    
    \includegraphics[width=0.18\linewidth]{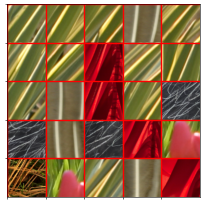}
    \includegraphics[width=0.22\linewidth]{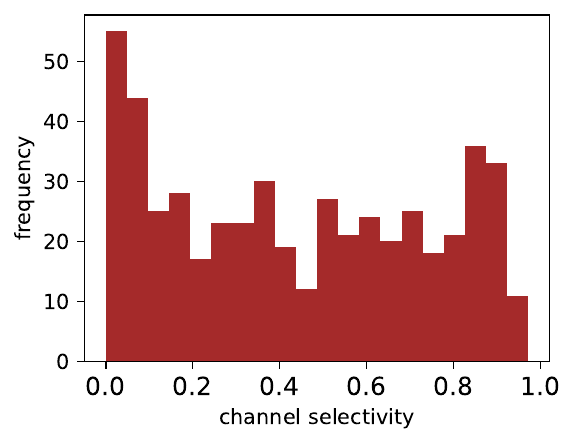}
    \hfil
    \includegraphics[width=0.18\linewidth]{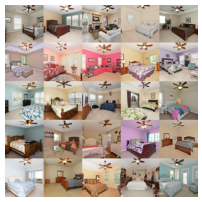}    
    \includegraphics[width=0.18\linewidth]{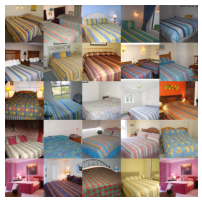}
    \includegraphics[width=0.18\linewidth]{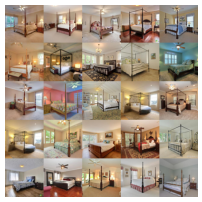}    
    \includegraphics[width=0.18\linewidth]{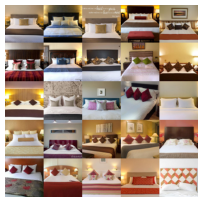}    
    \includegraphics[width=0.22\linewidth]{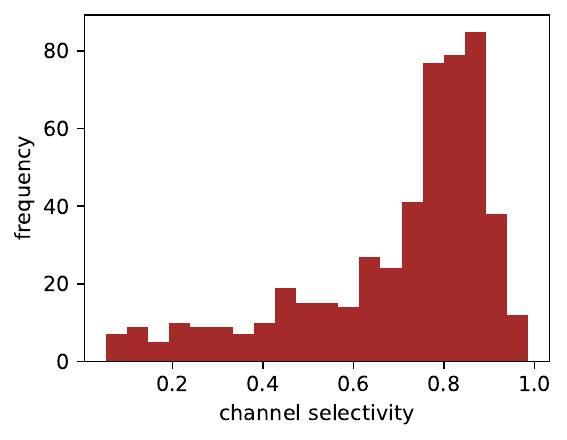}
    \hfil
    \includegraphics[width=0.18\linewidth]{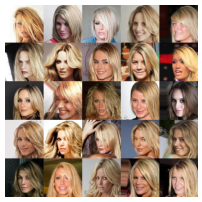}    
    \includegraphics[width=0.18\linewidth]{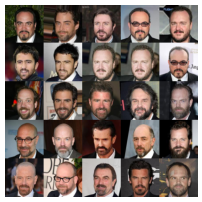}
    \includegraphics[width=0.18\linewidth]{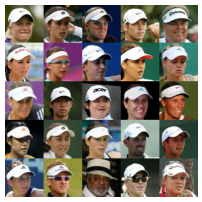}    
    \includegraphics[width=0.18\linewidth]{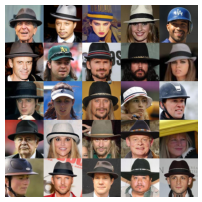}     
    \caption{{Channel selectivity for three models trained on {Texture}, {LSUN-Bedroom}, and {CelebA} datasets.} \textbf{Texture dataset:} Distribution of channel selectivity shows that the vast majority of channels have low PR, corresponding to channels that are highly specialized (and infrequently active), concentrated on the left. The panels show the set of images that maximally activate each of four specialized channels, revealing selectivity for different texture patterns. PR of channels from left to right for texture model: $0.013, 0.024, 0.027,0.038$.
    \textbf{LSUN-Bedroom:} Channel selectivity is distributed more evenly with more common channels compared to the models trained on ImageNet and Texture datasets. This can be attributed to the presence of more common patterns and structures within the images of the dataset, as they all depict bedrooms. The PR of the four channels shown are $0.015, 0.053, 0.018, 0.36,$, from left to right. 
    \textbf{CelebA:} Most channels in this model respond to the majority of images. This can be attributed to the fact that all images in the dataset share the global layout and coarser level structures since the faces are aligned and centered. The four channels shown here have PR of $ 0.29, 0.309, 0.36, 0.50$. 
    }
    \label{fig:channel-selectivity-texture}
\end{figure}

\begin{figure}[H]
    \centering
    \includegraphics[width=0.45\linewidth]{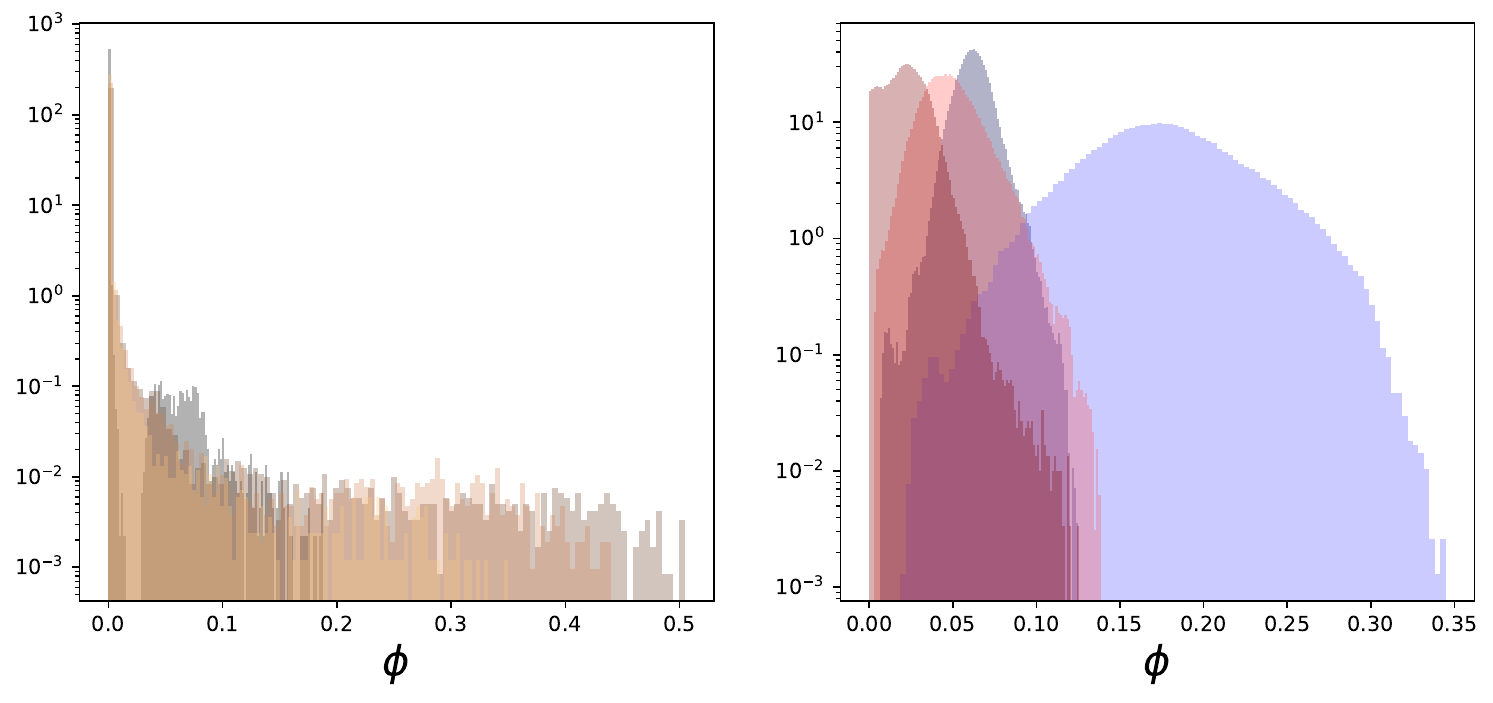}
    \includegraphics[width=0.26\linewidth]{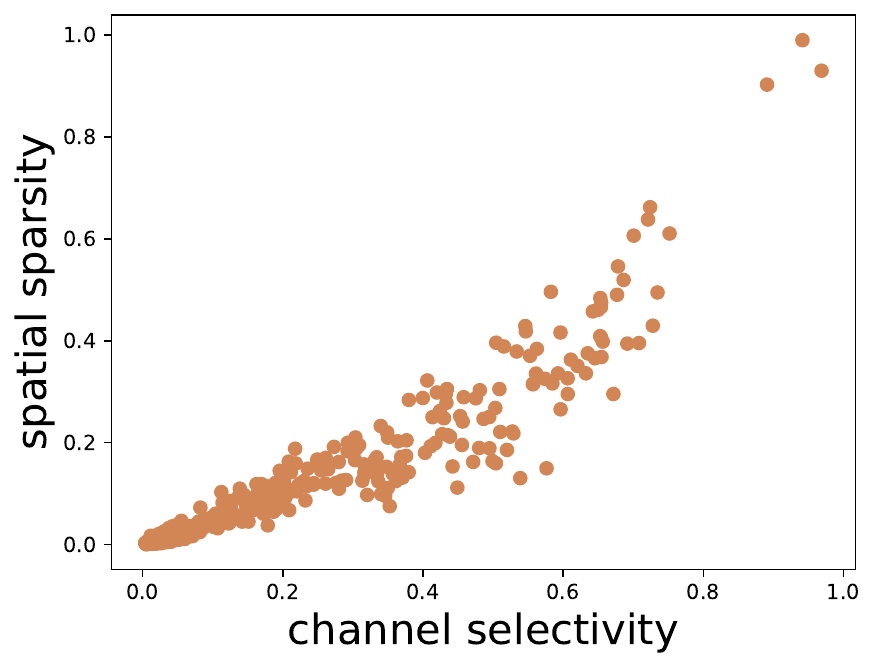}
    \includegraphics[width=0.26\linewidth]{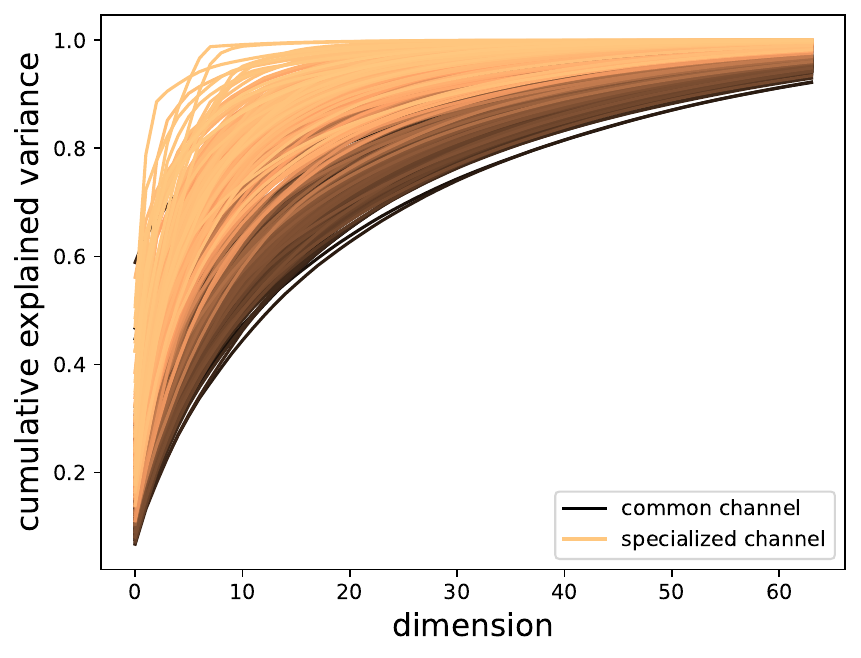}    
    \includegraphics[width=0.45\linewidth]{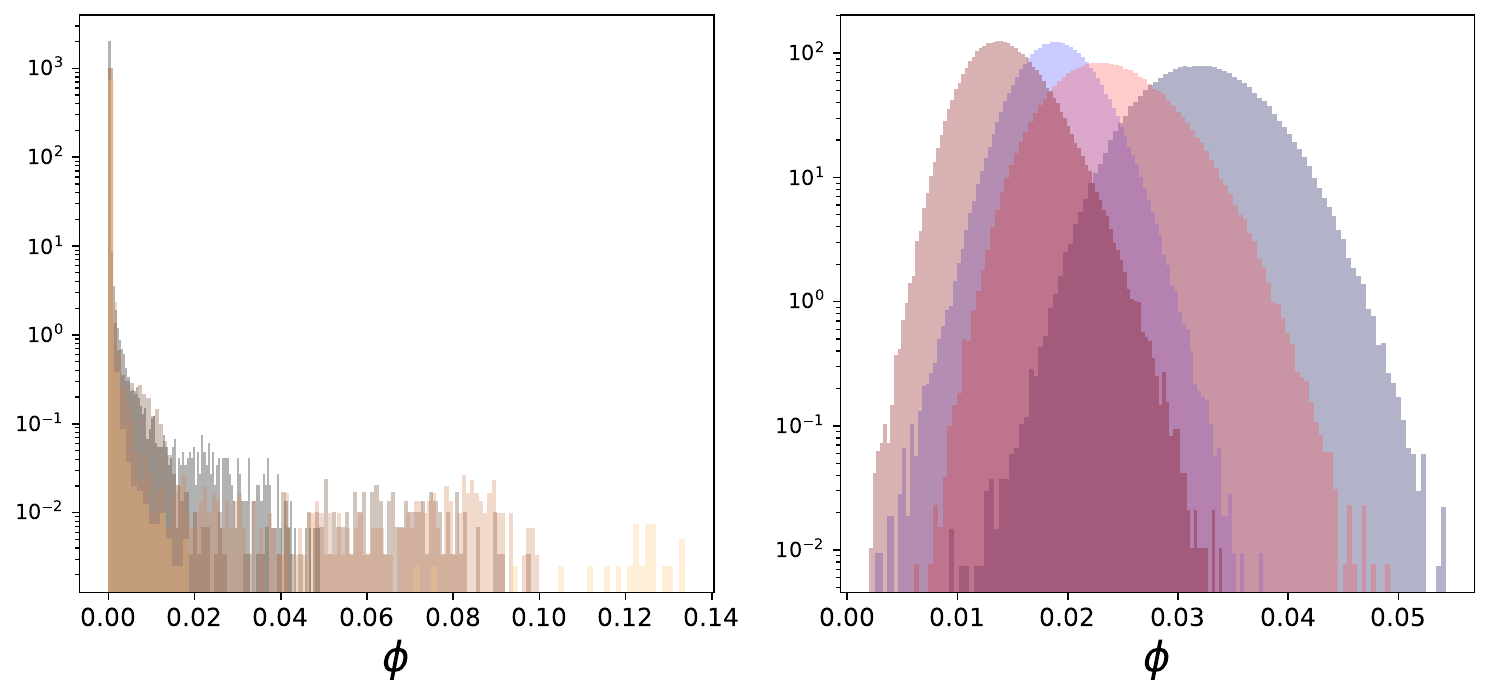}
    \includegraphics[width=0.26\linewidth]{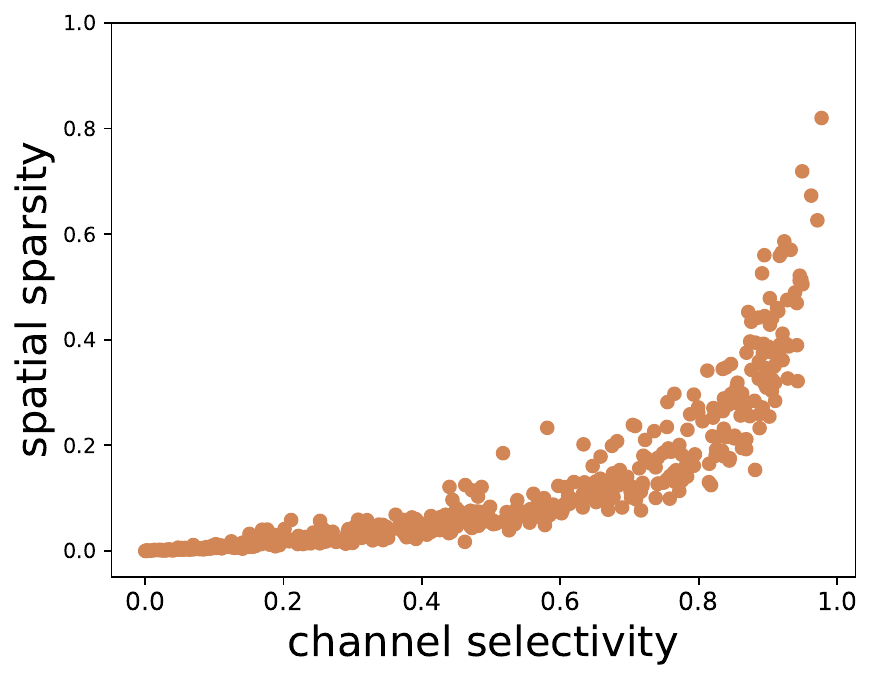}
    \includegraphics[width=0.26\linewidth]{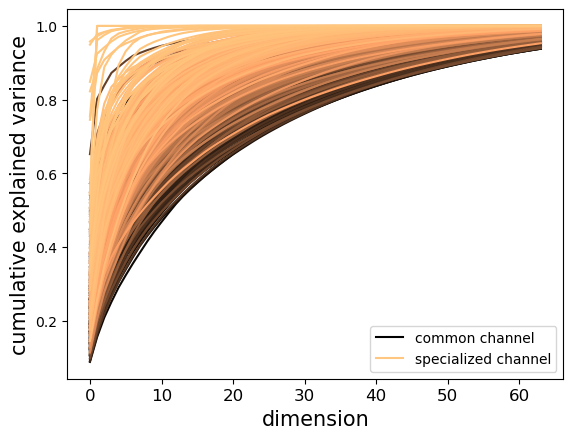}     
    \includegraphics[width=0.45\linewidth]{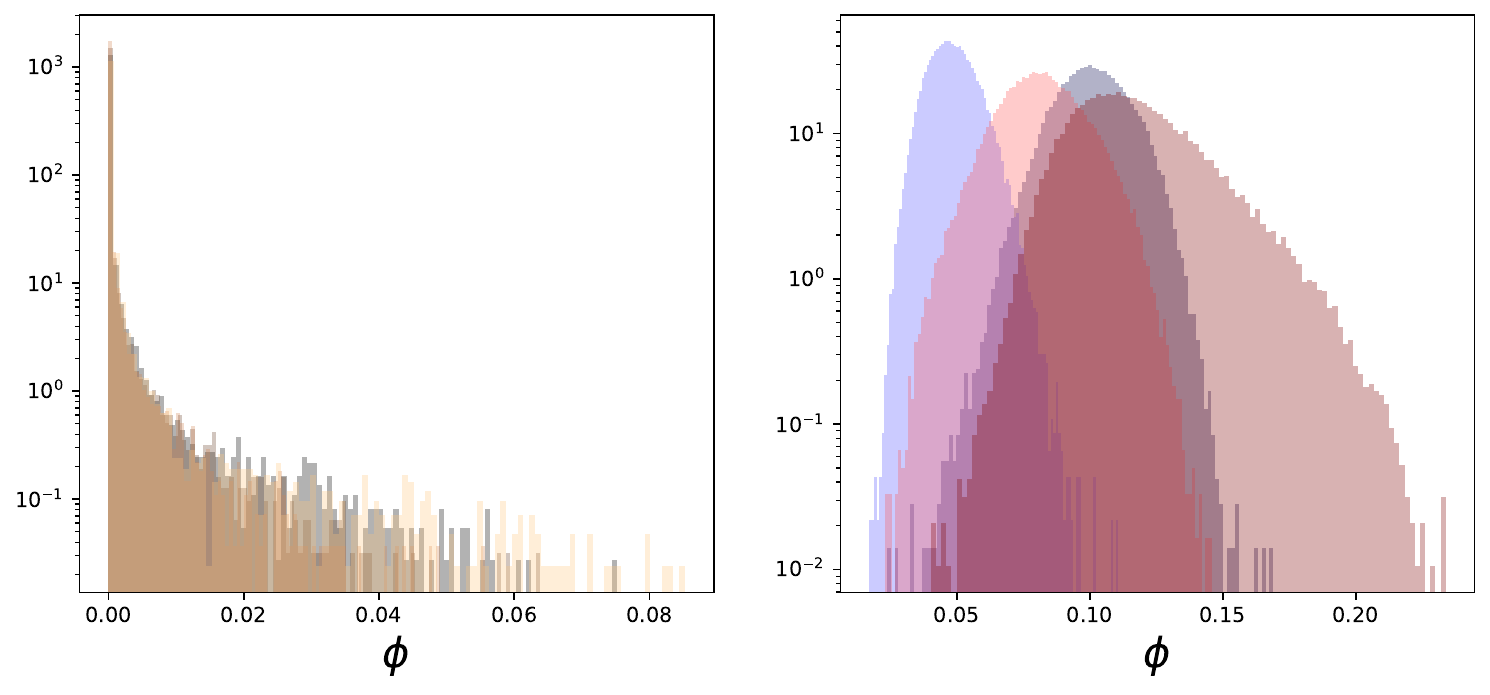}
    \includegraphics[width=0.26\linewidth]{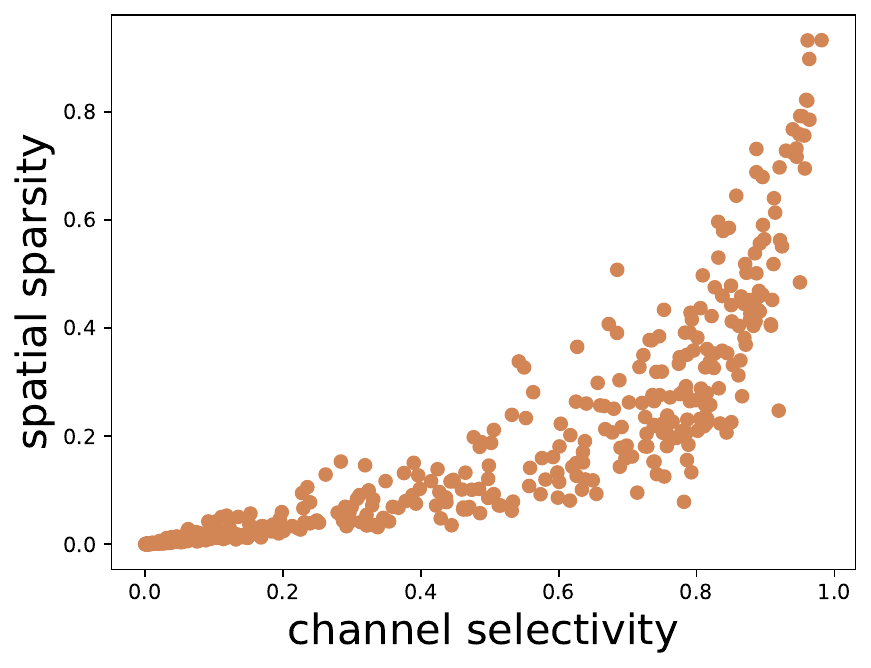}
    \includegraphics[width=0.26\linewidth]{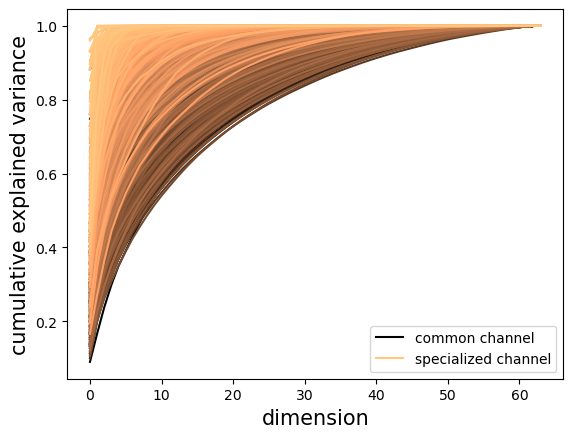}      
    \includegraphics[width=0.45\linewidth]{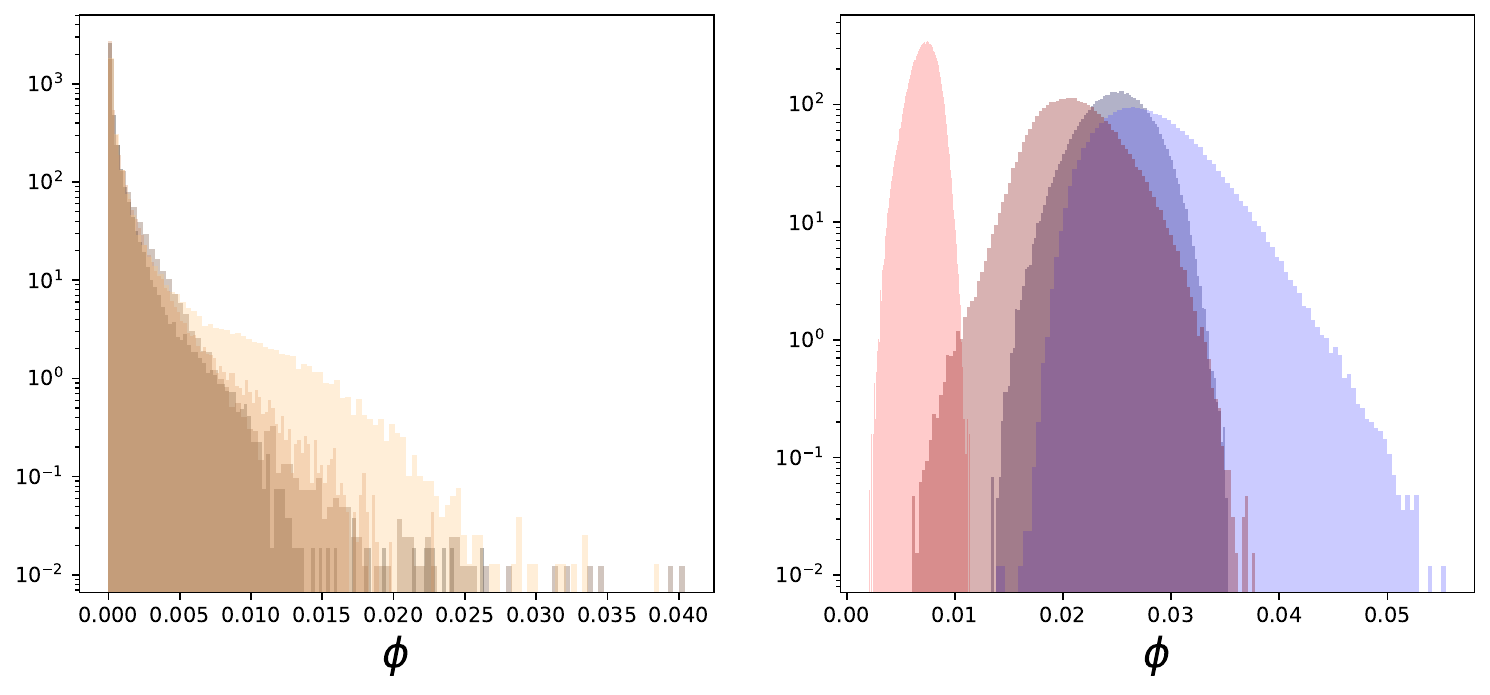}
    \includegraphics[width=0.26\linewidth]{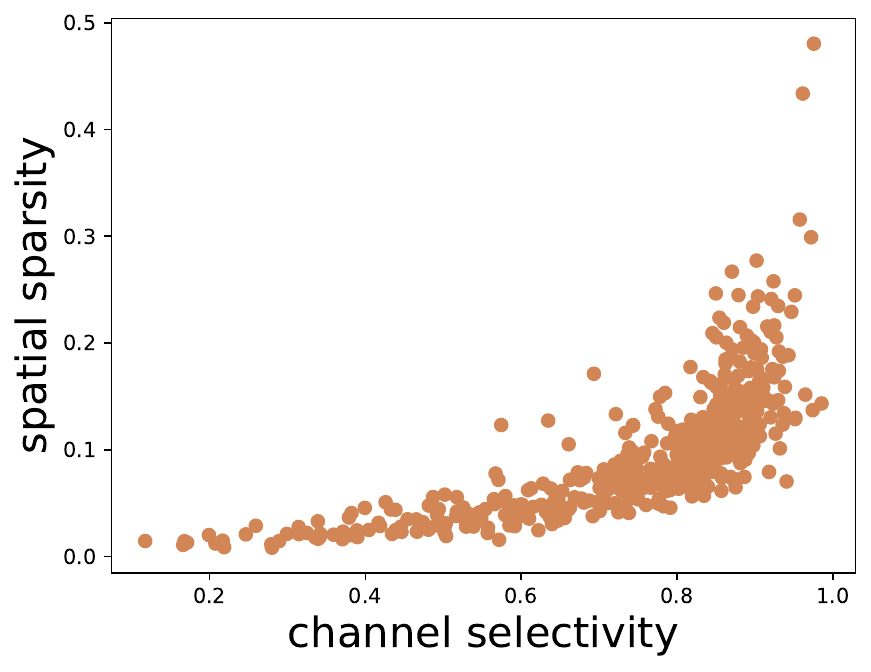}
    \includegraphics[width=0.26\linewidth]{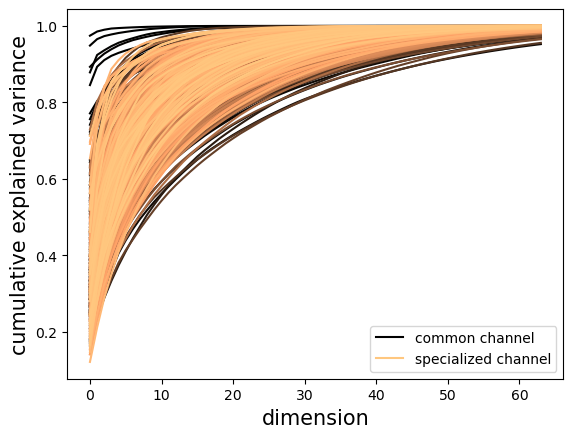}        
    \caption{{Channel selectivity predict signature statistical properties of the channels.} From top to bottom row, models trained on \textbf{Texture dataset}, \textbf{LSUN-bedroom}, \textbf{ImageNet} and \textbf{CelebA}. From left to right: 1) marginal distribution of $\phi[i]$ in 4 specialized channels. 2)  marginal distribution of $\phi[i]$ in 4 common channels. 3) channel selectivity is correlated with spatial sparsity within the channel. 4) The spatial variance of specialized channels is explained with fewer dimensions. 
    }
    \label{fig:statistics of channels}

\end{figure}
\subsection{Semantic similarities in representation space}

\begin{figure}[H]
    \centering
    \includegraphics[width=0.20\linewidth]{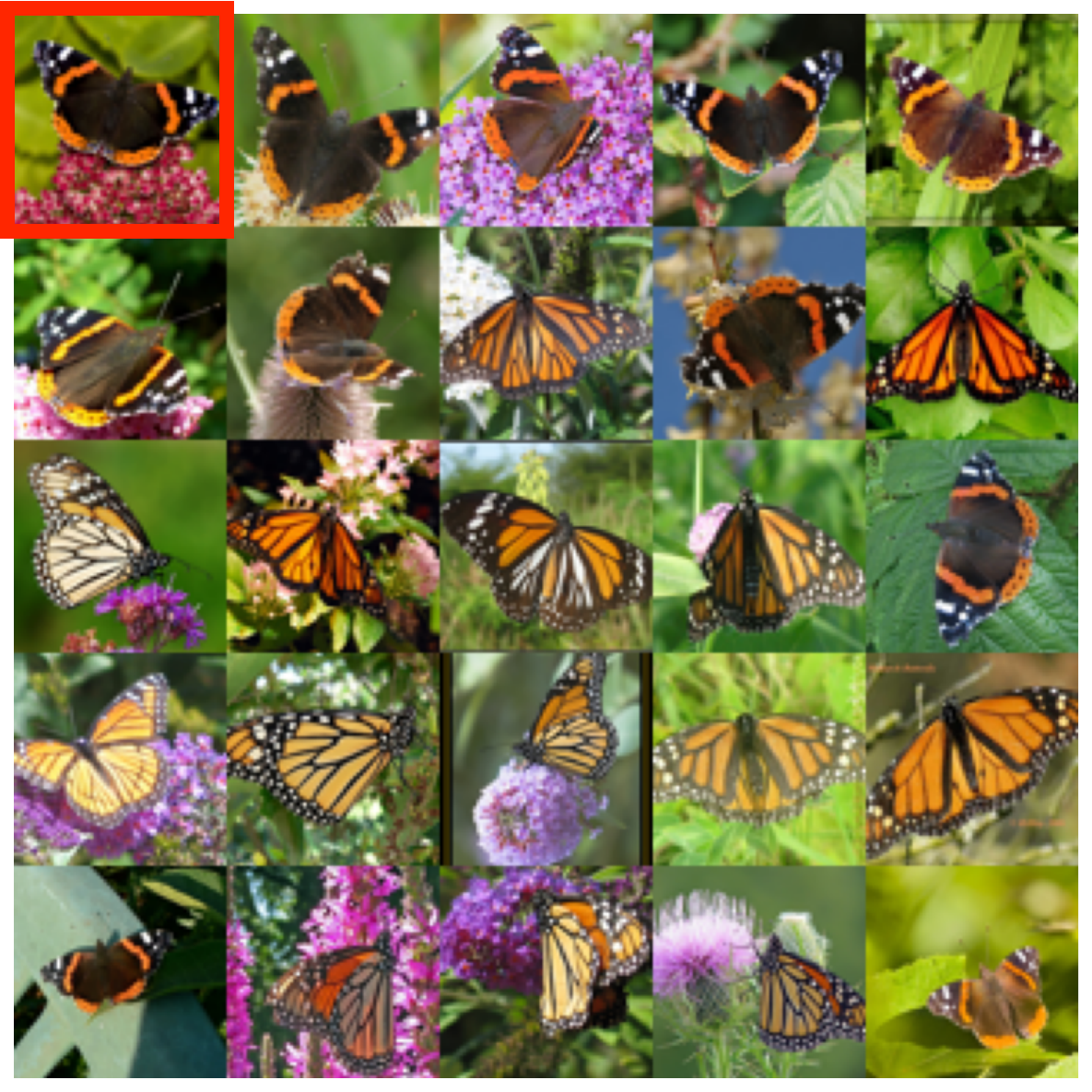}
    \includegraphics[width=0.20\linewidth]{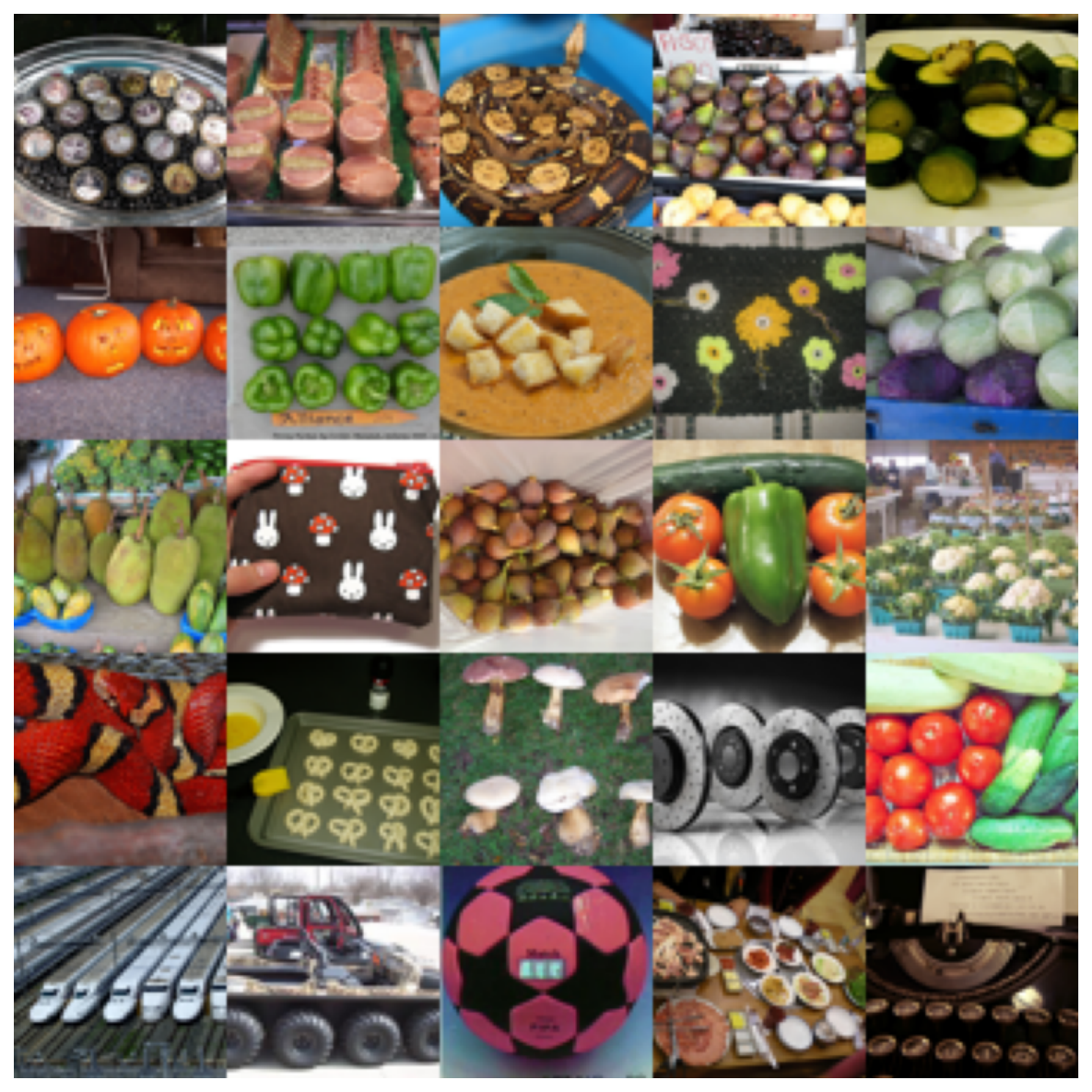}
    \hfil
    \includegraphics[width=0.20\linewidth]
    {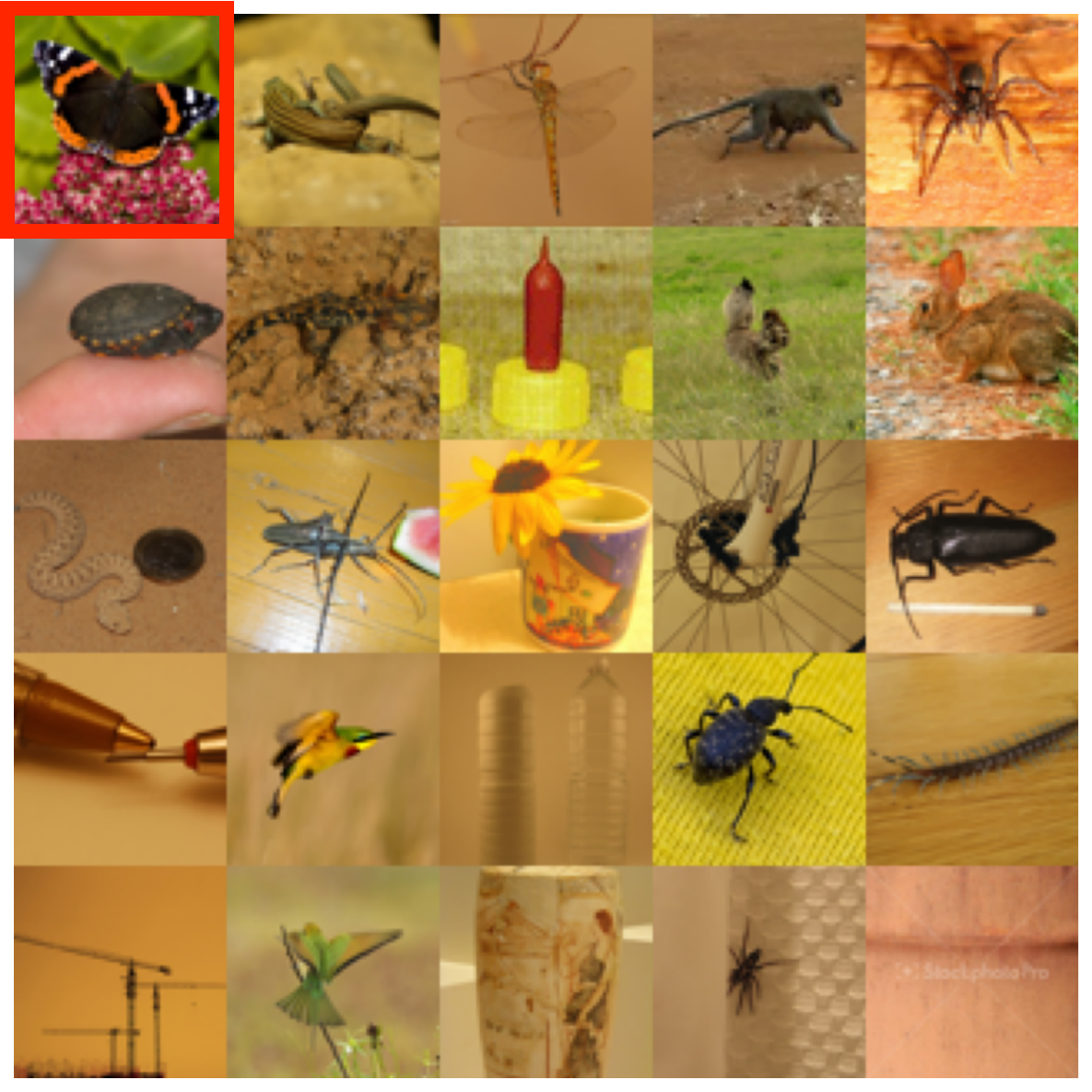}
    \includegraphics[width=0.20\linewidth]{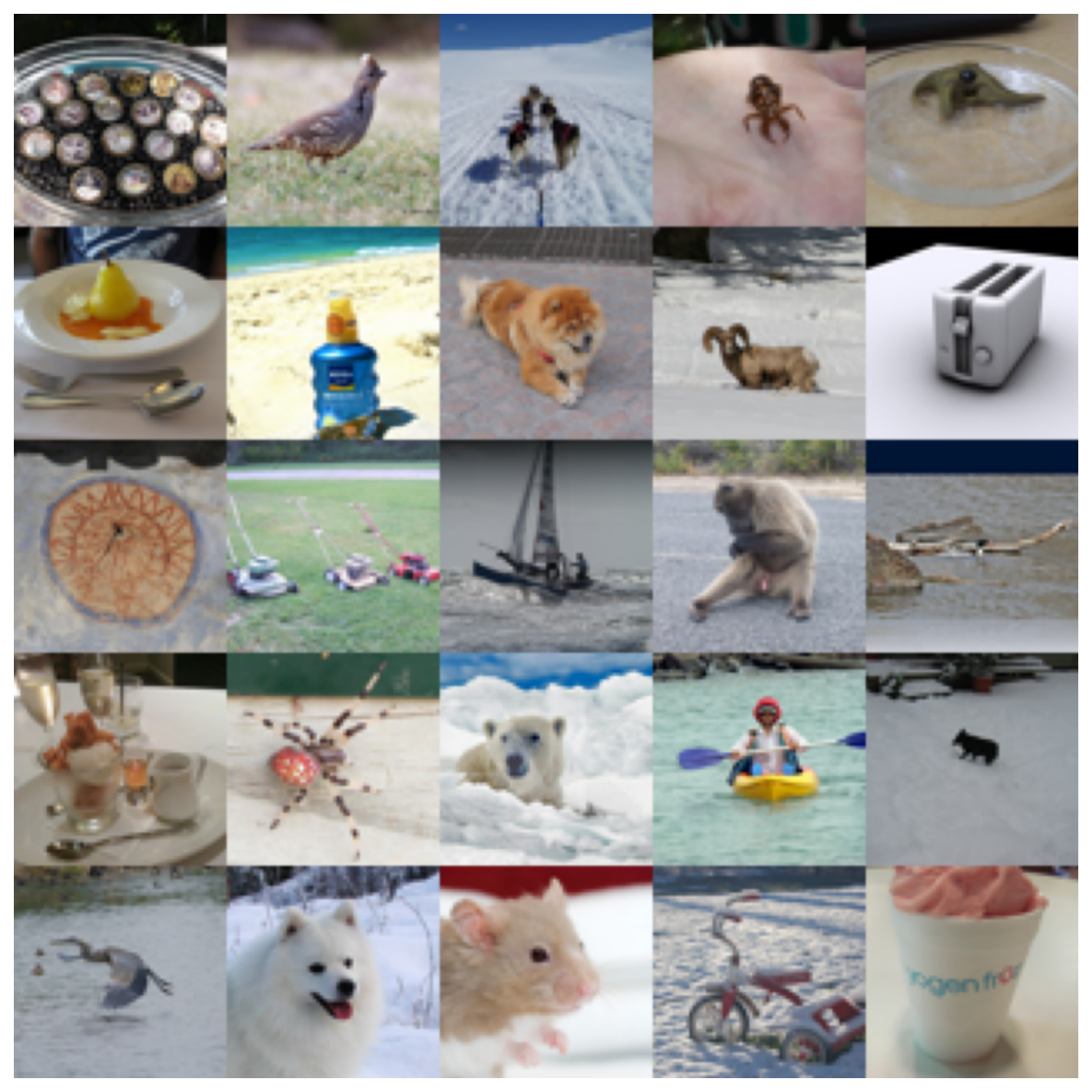} 

    \caption {{Nearby $\phi$'s correspond to images that are semantically similar.} \textbf{Left:} Two sets of images, each showing a target image (upper left), and a set of images whose $\phi$'s are closest to that of the target image in terms of cosine similarity.
    For the butterfly image, proximity in latent space is aligned with the object identity. For the tray of food, proximity is aligned with images depicting collections of similar objects. 
    \textbf{Middle:} Same as left, but showing images closest to the target in terms of cosine similarity in the pixel domain. \textbf{Right} Symmetrized KL divergence between the distributions of conditional samples, plotted against Euclidean distance in the representation, for randomly selected pairs of $\phi$ vectors. }
    \label{fig:distances-pixel-latent1}
\end{figure}

\paragraph{clustering.} We used the {\tt python} implementation of K-Means clustering algorithm, from the {\tt sklearn} package. 

Clustering at different noise levels leads to similar results, consistent with stability of $\phi$ over noise levels shown in \Cref{fig:noise-level-dependency-unet}. This is not true when noise is very small, since $\phi$ collapses to zero at small noise levels. 

K-mean clustering algorithms are sensitive to initialization, so the assignments of images to clusters changes with initialization, but always leads to the same semantic grouping. Even initialization from the centroids of the pre-defined class labels results in similar clusters. This means that even when we give the algorithm a chance to cluster the images within the same class together, it pushes away from that and re-allocate the assignments such that the images are grouped together based on "the gist of the scene" as opposed to object identity.

\begin{figure}[H]
    \centering
    \foreach \i in {6,11,32,34,45,35,37,38,39,43,44, 48,50, 42,21,31,17,25} {
        \includegraphics[width=1\textwidth]{figures/mixture_model/embedding/c-\i.png}
    }
    \caption{Continued from \Cref{fig:dprime-tSNE}. Random images from different clusters are shown in each row. Different cars with the same orientation are clustered together, but similar car brands with different orientations are assigned to different clusters. (see rows 2 and 3). }
    \label{fig:images in different clusters2}
\end{figure}

\begin{figure}[H]
    \centering
    \foreach \i in {40,113,154,223,236,297,370,425,520,535,585,642,764,915,981,1057,1114,1115,1143} {
        \includegraphics[width=1\textwidth]{figures/mixture_model/embedding/cluster_examples_texture_\i.png}
    }
    \caption{Random images from different clusters are shown in each row. Obtained from the model trained on \textbf{Texture dataset}.}
    \label{fig:images in different clusters - texture}
\end{figure}

%%%%%%%%%%%%%%%%%%%%%%%%%%%%%%%%%%%%%%%%%%%%%%%%%%%%%%%%%%%%%%%%%%%%%%%%%%
\section{Stochastic reconstruction algorithm}
\label{app:sampling results}
We build our sampling algorithm based on the algorithm in \citep{kadkhodaie2021stochastic} which does not require the noise level and follows an adaptive step size schedule. Hence, this network does not take the noise variance as an input and is a blind denoiser. This setup has the advantage that simplifies the network, which is important for analysis of the internal layers of the model.

\begin{figure}
    \centering
\includegraphics[width=0.19\linewidth]{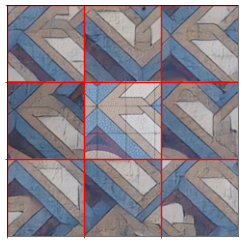}  
\includegraphics[width=0.19\linewidth]{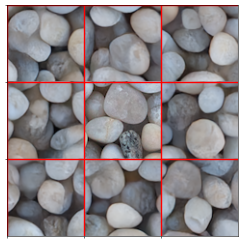} 
\includegraphics[width=0.19\linewidth]{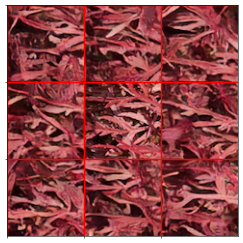} 
\includegraphics[width=0.19\linewidth]{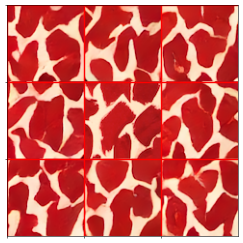} 
\includegraphics[width=0.19\linewidth]{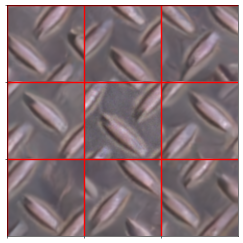} 
\includegraphics[width=0.19\linewidth]{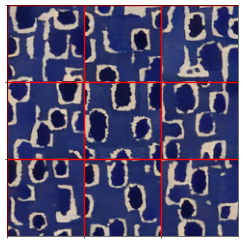}  
\includegraphics[width=0.19\linewidth]{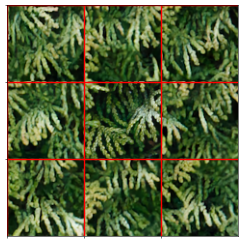} 
\includegraphics[width=0.19\linewidth]{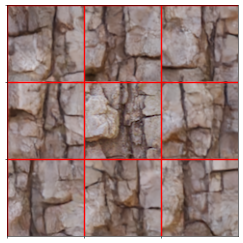} 
\includegraphics[width=0.19\linewidth]{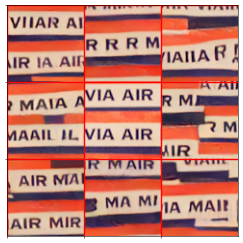} 
\includegraphics[width=0.19\linewidth]{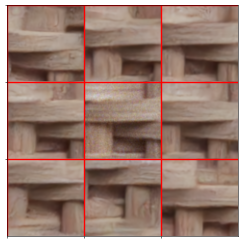} 
\includegraphics[width=0.19\linewidth]{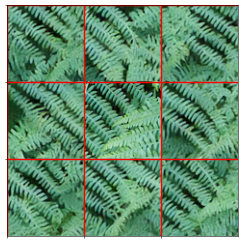}  
\includegraphics[width=0.19\linewidth]{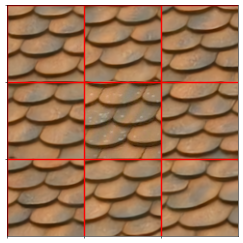} 
\includegraphics[width=0.19\linewidth]{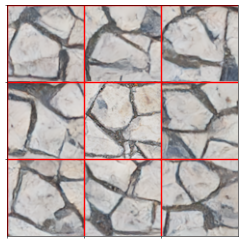} 
\includegraphics[width=0.19\linewidth]{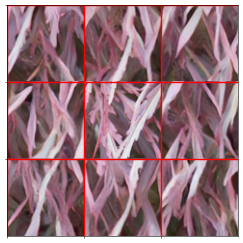} 
\includegraphics[width=0.19\linewidth]{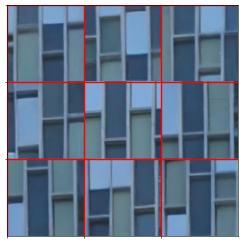}     
\caption{More samples from model trained on Texture dataset. See caption of \Cref{fig:samples-imagenet}. }
\label{fig:sampes-textures}
\end{figure}

\begin{figure}
    \centering
    \includegraphics[width=0.195\linewidth]{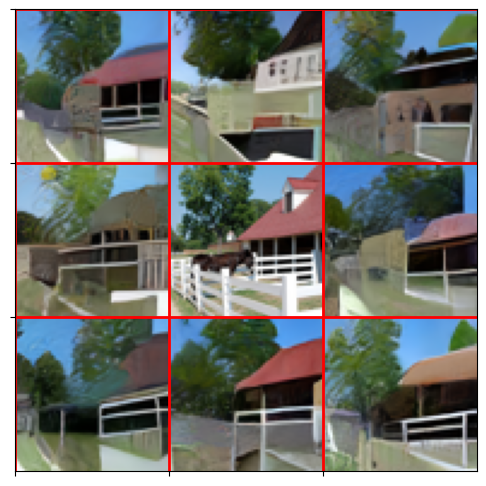}
    \includegraphics[width=0.195\linewidth]{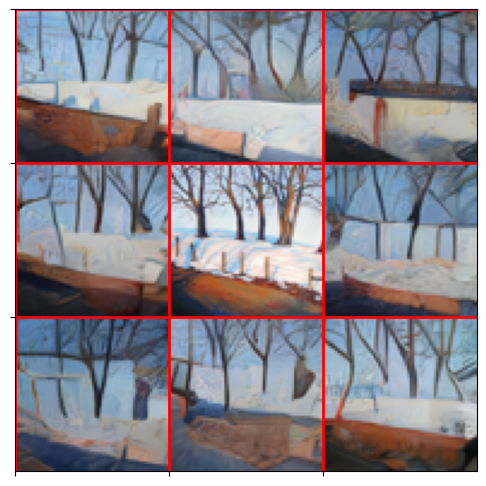}       
    \includegraphics[width=0.195\linewidth]{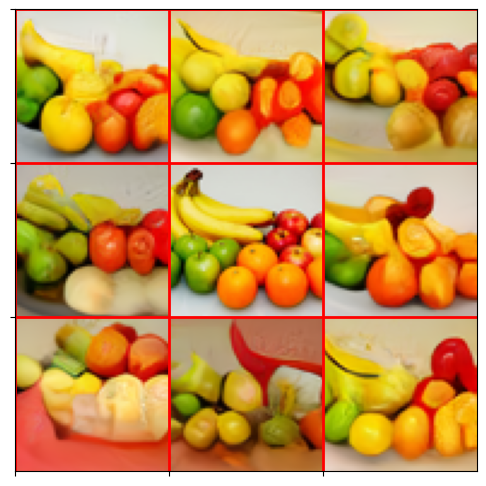}     
    \includegraphics[width=0.195\linewidth]{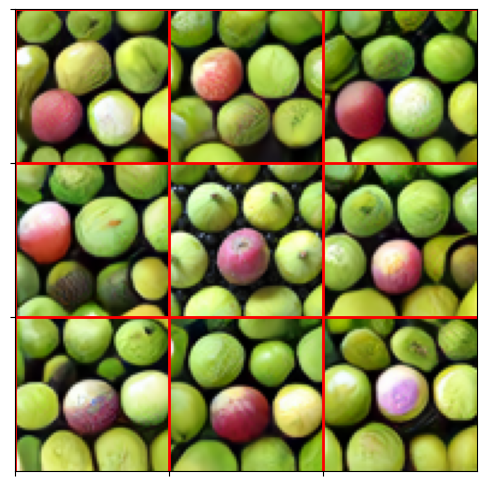}    
    \includegraphics[width=0.195\linewidth]{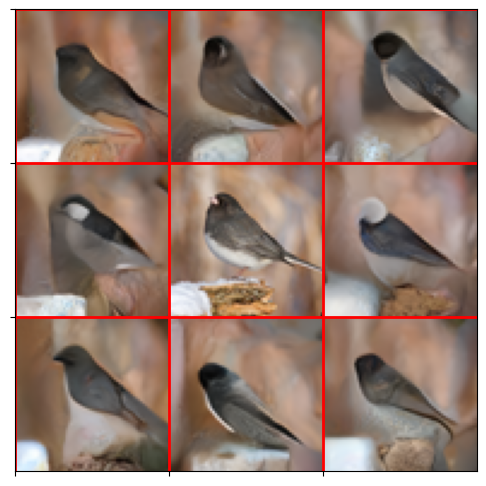}      
    \includegraphics[width=0.195\linewidth]{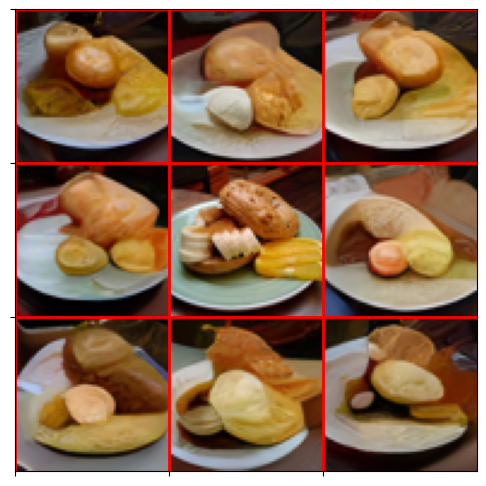} 
    \includegraphics[width=0.195\linewidth]{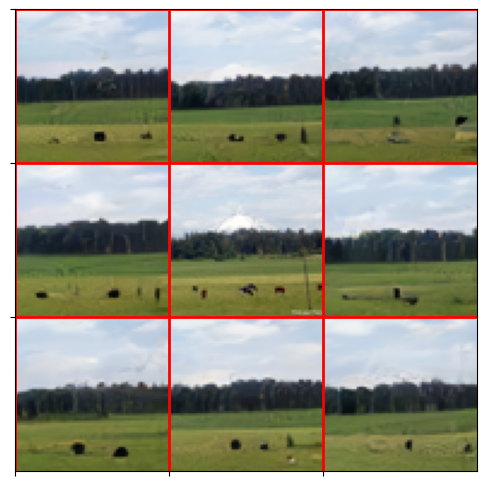}   
    \includegraphics[width=0.195\linewidth]{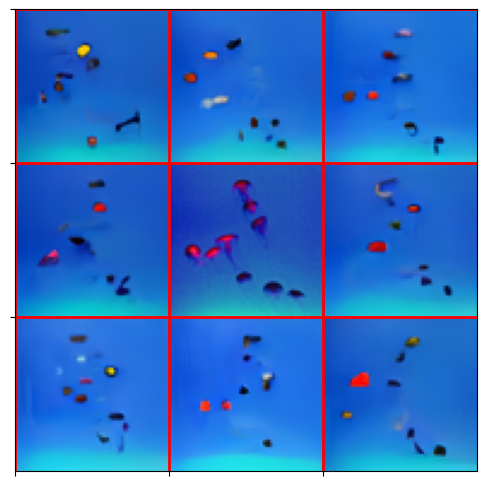}     
    \includegraphics[width=0.195\linewidth]{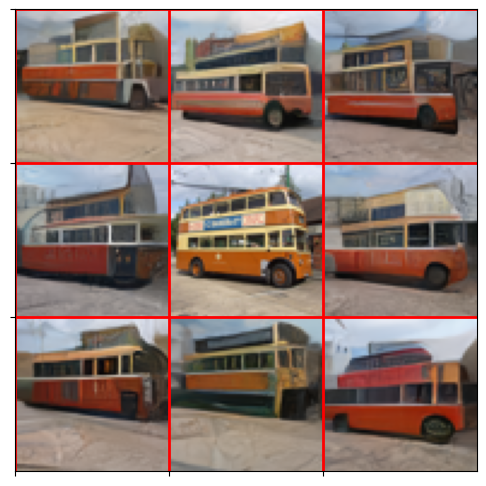}
    \includegraphics[width=0.195\linewidth]{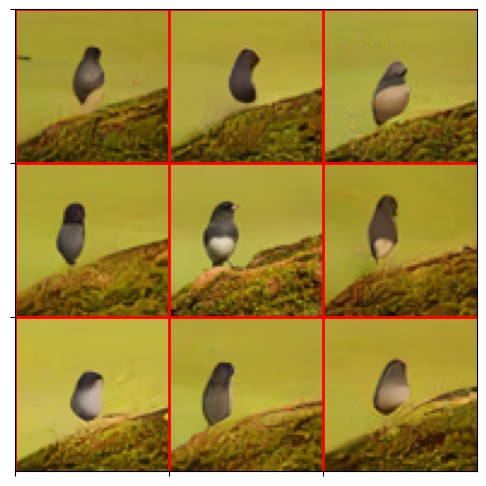}     
    \includegraphics[width=0.195\linewidth]{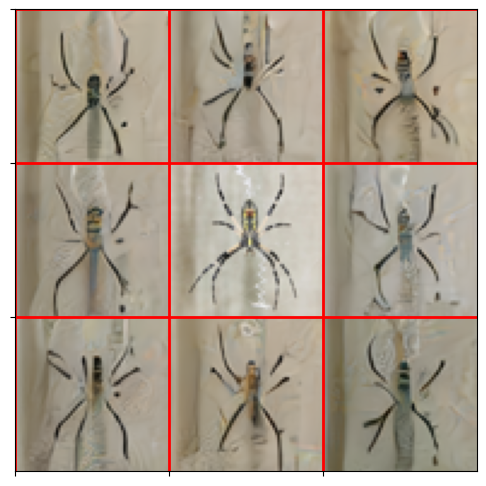} 
    \includegraphics[width=0.195\linewidth]{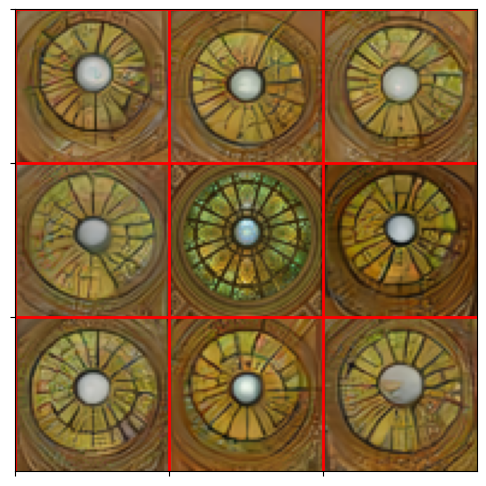}
    \includegraphics[width=0.195\linewidth]{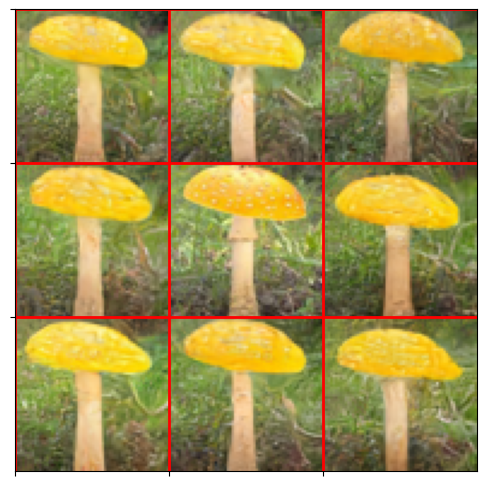}   
    \includegraphics[width=0.195\linewidth]{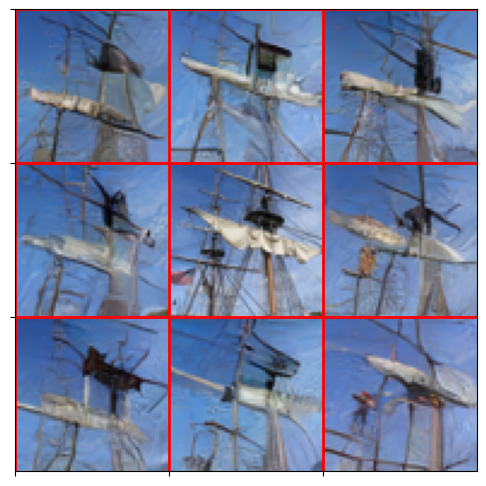}   
    \includegraphics[width=0.195\linewidth]{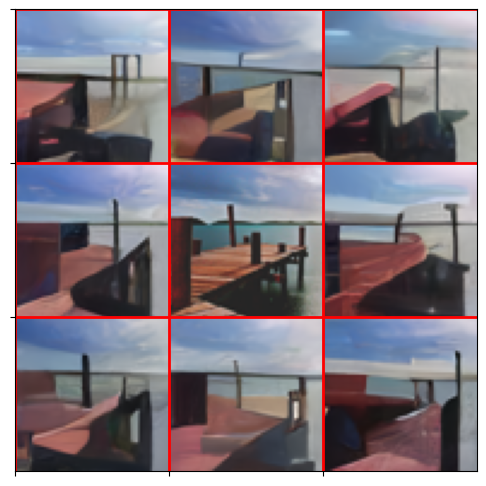}     
    \caption{More samples from model trained on ImageNet64 dataset. The samples are visually similar to the target image at the center of each panel. Interestingly, the location of the large, long-ranging image patterns are persevered in the samples, while the fine structures and details are diverse in their location with respect to the target image. 
    See caption of \Cref{fig:samples-imagenet}.}
    \label{fig:samples-imagenet2}
\end{figure}

\begin{figure}
    \centering
\includegraphics[width=0.22\linewidth]{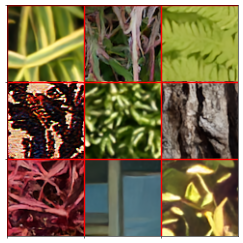}  
\includegraphics[width=0.22\linewidth]{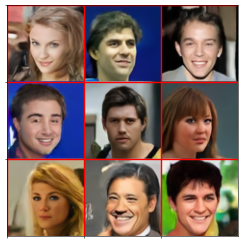} 
\includegraphics[width=0.22\linewidth]{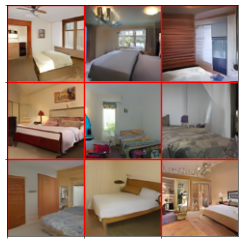} 
\includegraphics[width=0.22\linewidth]{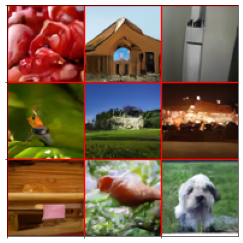} 
    \caption{Samples generated unconditionally from models trained on Texture, CelebA, LSUN-Bedrooms, and ImageNet64 datasets respectively. }
    \label{fig:unconditional_samples}
\end{figure}

% \begin{figure}[H]
%     \centering
%     \includegraphics[width=.7\linewidth]{figures/mixture_model/sampling/guidance_effect_low_res.pdf}    
%     \caption{A block diagram illustrating the matching step implementation and its effect. \textbf{Top:} One forward pass through the UNet without matching guidance. \textbf{Bottom:} One forward pass after matching representations in the middle block. The noise initialization, $x_t$, are identical in both passes, but the output of the network is vastly different for them, due to the effect of matching. Matching helps the large structures in the conditioner image appear in the output image. See more examples of this effect in \Cref{fig:effect-of-guaidance-exmaples}.}
%     \label{fig:effect-of-guaidance-digram}
% \end{figure}

\begin{figure}
    \centering
    \includegraphics[width=0.4\linewidth]{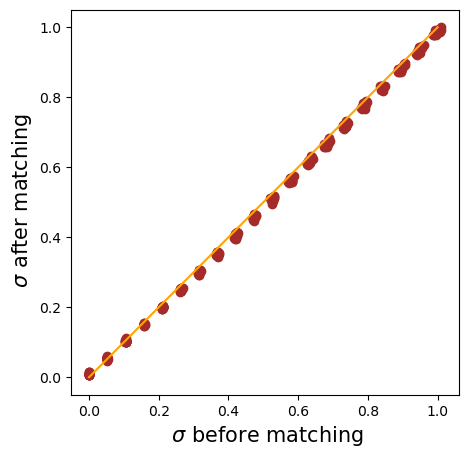}
    \caption{Matching does not change the noise level $\sigma$. This plots shows the standard deviation of noise on the image before and after the matching step, for a collection of random test images, at different input noise levels. $\sigma$ after matching is measured by the norm of score divided by the square root of ambient dimensionality, ${\Vert s(x_\sigma)\Vert}/{\sqrt{n}}$. This is an approximation of the strength of the remainder noise on the image by assuming a Gaussian prior. }
    \label{fig:noise-change-matching}
\end{figure}

\begin{figure}
    \centering
    \includegraphics[width=0.305\linewidth]{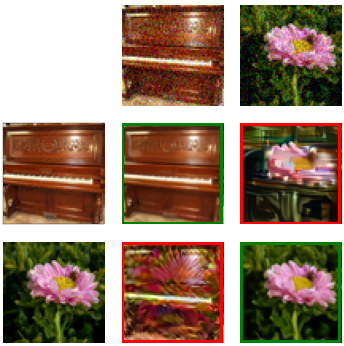}  
    \hspace{.5cm}
    \includegraphics[width=0.2\linewidth]{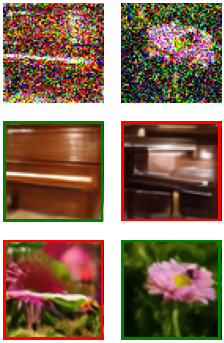}  
    \hspace{.5cm}
    \includegraphics[width=0.2\linewidth]{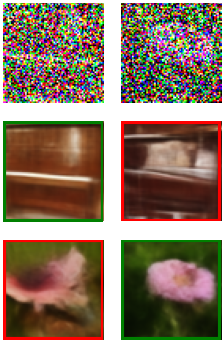}      
    % \hspace*{-1ex}
    \caption {A pair of clean images, $x$ and $x'$, from ImageNet dataset is shown on the first column. The distance between the the two densities induced by the $\phi$'s obtained from these two images can be defined and computed using \Cref{eq:density-distance}. The first term in the integrand is the difference between two conditional scores: the score of $x_{\sigma}$ given $\phi$ and the score of $x_{\sigma}$ given $\phi'$. This difference is computed and integrated at all $\sigma$ levels (and symmetrized). For these two images, this difference is visualized in the green versus red boxes. 
    }
    \label{fig:effect-of-guaidance-exmaples}
\end{figure}

%%%%%%%%%%%%%%%%%%%%%%%%%%%%%%%%%%%%%%%%%%%%%%%%%%%%%%%%%%%%
%%%%%%%%%%%%%%%%%%%%%%%%%%%%%%%%%%%%%%%%%%%%%%%%%%%%%%%%%%%%

\end{document}